\documentclass[fleqn,10pt]{wlscirep}

\usepackage[utf8]{inputenc}
\usepackage[T1]{fontenc}
\usepackage[cal=cm]{mathalfa}
\usepackage{graphicx}
\usepackage{subcaption}  % For subfigures
\usepackage{booktabs} % for professional tables
\usepackage{multirow}
\usepackage{hyperref}

\title{Temporal Neural Operator for Modeling Time-Dependent Physical Phenomena}

\author[1]{Waleed Diab}
\author[1,2,*]{Mohammed Al Kobaisi}

\affil[1]{Khalifa University, Chemical and Petroleum Engineering, Abu Dhabi, UAE}
\affil[2]{Delft University of Technology, Delft Institute of Applied Mathematics, Delft, The Netherlands}

\affil[*]{M.S.K.AlKobaisi@tudelft.nl; mohammed.alkobaisi@ku.ac.ae}


\begin{document}

\begin{abstract}
Neural Operators (NOs) are machine learning models designed to solve partial differential equations (PDEs) by learning to map between function spaces. Neural Operators such as the Deep Operator Network (DeepONet) and the Fourier Neural Operator (FNO) have demonstrated excellent generalization properties when mapping between spatial function spaces. However, they struggle in mapping the temporal dynamics of time-dependent PDEs, especially for time steps not explicitly seen during training. This limits their temporal accuracy as they do not leverage these dynamics in the training process. In addition, most NOs tend to be prohibitively costly to train, especially for higher-dimensional PDEs. In this paper, we propose the Temporal Neural Operator (TNO), an efficient neural operator specifically designed for spatio-temporal operator learning for time-dependent PDEs. TNO achieves this by introducing a temporal-branch to the DeepONet framework, leveraging the best architectural design choices from several other NOs, and a combination of training strategies including Markov assumption, teacher forcing, temporal bundling, and the flexibility to condition the output on the current state or past states. Through extensive benchmarking and an ablation study on a diverse set of example problems we demonstrate the TNO long range temporal extrapolation capabilities, robustness to error accumulation, resolution invariance, and flexibility to handle multiple input functions.  
\end{abstract}

\flushbottom
\maketitle
\thispagestyle{empty}

\section*{Introduction}
Time-dependent Partial Differential Equations (PDEs) are fundamental tools for modeling the dynamic behavior of complex physical, biological, and environmental systems, including fluid dynamics, heat transfer, wave propagation, and biological processes. Despite their widespread applicability, significant challenges hinder their use in real-world scenarios. Many problems lack analytical solutions, necessitating computationally intensive numerical simulations. Moreover, incomplete or uncertain knowledge of the underlying physics can lead to inaccuracies in model formulation. In domains such as climate and weather forecasting, the availability of vast observational data presents both an opportunity and a challenge: integrating large-scale heterogeneous datasets—often marred by noise, gaps, and inconsistencies—into high-fidelity simulations requires advanced data assimilation techniques and substantial computational resources. These limitations underscore the need for alternative approaches that can effectively leverage available data while mitigating computational and modeling burdens.

In recent years, machine learning for scientific computing applications has gained wide spread popularity due to its high potency in handling large volumes of data. Physics-Informed Neural Networks (PINNs)\cite{RAISSI2019686} have been introduced as a deep learning framework that can learn nonlinear dynamics by incorporating physical laws, expressed as partial differential equations (PDEs), directly into the loss function; thus ensuring that the model adheres to the governing physical principles while fitting observed data. However, PINNs have not shown significant computational gains compared to traditional numerical methods, particularly when dealing with large-scale simulations or stiff systems, where the iterative nature of training deep neural networks can result in slower run-times and higher resource consumption. Neural operator learning, which aims to learn mappings between function spaces rather than individual point-wise solutions, was later introduced through the Deep Operator Network (DeepONet)\cite{Lu2021LearningOperators}. This approach enables models to generalize more effectively across varying inputs and provides a significant advantage in handling diverse physical systems. Architectures like the DeepONet\cite{Lu2021LearningOperators} and the Fourier Neural Operator (FNO)\cite{li2021FNO} exemplify this approach. These architectures have demonstrated superior scalability and computational efficiency when solving complex partial differential equations, especially in large-scale and high-dimensional problems.

While neural operators\cite{Lu2021LearningOperators,li2021FNO, Wen2022U-FNOAnFlow, Diab2023U-DeepONet:Sequestration, jiang2024fourier, lee2024efficient} provide a promising approach to address the challenges of solving time-dependent PDEs, they often struggle with extrapolation beyond the temporal training horizon. This limitation hampers their effectiveness for long-term predictions, as performance typically degrades when forecasting solutions beyond the temporal range of the training set. Moreover, these architectures often rely on fixed spatio-temporal grids which further reduces their flexibility in handling variable resolutions. 

In this work, we introduce the Temporal Neural Operator (TNO), a novel neural operator architecture that enables simultaneous generalization to new PDE instances and temporal extrapolation beyond the training dataset with negligible error accumulation over time. Furthermore, we leverage the TNO's generalization capabilities to learn 3D problems using 2D operations, which leads to significant reductions in training cost. This feature is demonstrated in our first example of a 3D global daily air temperature prediction, along with long-term temporal extrapolation. In the second example, we showcase the TNO's zero-shot super-resolution capability using historical daily observational air temperatures over Europe. Here, the TNO is trained on historical air temperature data at a 0.25° grid resolution, and tested on future air temperature data at a 0.1° grid resolution. The TNO demonstrates superb performance in the simultaneous temporal extrapolation and super-resolution with high accuracy and error invariance to resolution. In the third example, we validate the TNO's ability to generalize across diverse input functions and variables on a carbon sequestration dataset which tracks both saturation and pressure buildup. Despite the limited number of available time steps, the diversity of PDE variables, and the complexity of the problem, the TNO achieves robust temporal extrapolation and generalization performance. The three examples presented herein encapsulate the versatile potential of TNO to tackle complex, time-dependent real-world problems. Overall, we make the following key contributions:
\begin{itemize}
    \item Architecture: We introduce the temporal-branch (t-branch), encoder layers and U-Net blocks in the branch and t-branch, we replace the dot product with the Hadamard product, and introduce a decoder Feed-Forward Neural Network (FFN).
    \item Training: We devise a flexible training strategy with temporal-bundling to encourage stable long rollouts into the future and reduce error accumulation, as well as teacher forcing for accelerated training.
    \item Flexibility: The TNO is a flexible framework which can be conditioned autoregressively or on a memory of past states.  
    \item Efficiency: We demonstrate how to utilize the strong generalization capabilities of the TNO to model 3D problems using 2D operations.   
\end{itemize}
The combination of strong generalization to multiple new PDE parameters, long accurate temporal extrapolation with minimal error accumulation, flexible input and output step size, in addition to the TNO low memory footprint, and super resolution capabilities, positions the TNO as the State-Of-The-Art in neural operator learning for time dependent PDEs. To strengthen this claim, the TNO capabilities are demonstrated on three real-world challenging problems.

\subsection*{Time-Dependent Partial Differential Equations}
Time-dependent partial differential equations (PDEs) describe the spatio-temporal evolution of a solution, denoted as $u(t, \mathbf{x})$, across a temporal domain $t \in [0, T]$ and a spatial domain $\mathbf{x} = [x_1, x_2, ..., x_m] \in \mathcal{X} \subseteq \mathbb{R}^m$. These PDEs relate the temporal derivative $u_t$ to the spatial derivatives $u_{\mathbf{x}}$, $u_{\mathbf{xx}, \dots}$,    through a general functional form $F$, such that:

\begin{equation}
    u_t = F(t, \mathbf{x}, u, u_{\mathbf{x}}, u_{\mathbf{xx}}, ...).
    \label{eq:time-PDE}
\end{equation}

One way for analyzing the temporal dynamics of these systems is offered by operator learning, which focuses on identifying mappings between function spaces. In this context, we define a time evolution operator $\mathcal{G}_{\Delta t} : \mathbb{R}_{> 0} \times \mathbb{R}^n \rightarrow \mathbb{R}^n$ that directly propagates the solution forward in time. Specifically, the solution at a future time $t + \Delta t$ is obtained by applying this operator to the spatial profile of the solution at the current time $t$, denoted as $u(t, \cdot)$:

\begin{equation}
    u(t + \Delta t) = \mathcal{G}_{\Delta t}(t, u(t, \cdot)).
    \label{time-operator}
\end{equation}

The operator $\mathcal{G}$ acts on function spaces, mapping $u \in \mathcal{U}$ to $ u' \in \mathcal{U}'$, where the input function $u \in \mathcal{U}$ is mapped to the output function $u(t + \Delta t) \in \mathcal{U}'$. Here, $u : \mathcal{X} \rightarrow \mathbb{R}^n$, $u' : \mathcal{X}' \rightarrow \mathbb{R}^{n'}$, and the spatial domain is defined such that $\mathcal{X} \in \mathbb{R}^{m}$, and $\mathcal{X}' \in \mathbb{R}^{m'}$. Equation \ref{time-operator} implicitly assumes a Markov property for neural operators, where future state of the system depends only on the current state and not on the history of past states. As such, we say that the neural operator is autoregressively parametrized. Alternatively, the neural operator can be conditioned on the memory of past system states. Consequently, equation \ref{time-operator} can be extended to an arbitrary number of input time steps $L$ and output time steps $K$, where $K$ is the temporally bundled predictions\cite{Brandstetter2022MessageSolvers}.

Extending the single-step formulation, we define a history of $L$ past states of the solution as
\begin{equation}
    \mathbf{U}_{hist}(t) = \{u(t - (l-1)\Delta t, \cdot)\}_{l=1}^{L},
    \label{U_hist}
\end{equation}

and a bundle of $K$ future states as 

\begin{equation}
    \mathbf{U}_{fut}(t) = \{u(t + k\Delta t, \cdot)\}_{k=1}^{K}.
    \label{U_fut}
\end{equation}

The extended time evolution operator, denoted by $\mathcal{G}_{\Delta t}^{L \rightarrow K}$, maps the input history to the sequence of future states:

\begin{equation}
    \mathbf{U}_{fut}(t) = \mathcal{G}_{\Delta t}^{L \rightarrow K}(t, \mathbf{U}_{hist}(t)).
    \label{time-operator_K_L}
\end{equation}

More explicitly, this can be written as:

$$\begin{pmatrix}
u(t + \Delta t, \cdot) \\
u(t + 2\Delta t, \cdot) \\
\vdots \\
u(t + K\Delta t, \cdot)
\end{pmatrix} = \mathcal{G}_{\Delta t}^{L \rightarrow K}\left(t, \begin{pmatrix}
u(t - (L-1)\Delta t, \cdot) \\
\vdots \\
u(t - \Delta t, \cdot) \\
u(t, \cdot)
\end{pmatrix}\right).$$

Many architectures \cite{michalowska2024neural, ruiz2024benefits, he2024sequential, hu5149007deepomamba, hu2024state} approach neural operator learning of time dependent PDEs by utilizing a memory module to keep a compressed memory of the system past states, which is akin to choosing $L > 1$. On the other hand, choosing $L = 1$ regains the autoregressive approach found in many other architectures \cite{tran2021factorized, jiang2024fourier, lee2024efficient}. Choosing $L = 1$ or $L > 1$ is problem specific, however, we found that it is always a good idea to choose $K > 1$ as it reduces the number of solution calls. Moreover, $K = K_{max}$, that is, predicting all available time steps at once may cause a neural operator to lose its ability to learn temporal dynamics, and subsequently the temporal extrapolation ability. 

The TNO learns the time evolution operator from the data and effectively predicts the future states of the system governed by the PDE without explicit knowledge of the underlying differential equation $F$. This data-driven approach provides a direct pathway for forecasting the temporal behavior of PDE solutions based on observed or computed past states. In theory, a well-trained model can be rolled-out indefinitely; in practice however, error accumulation with excessive successive roll-outs render most temporal neural operators useless after a relatively short time horizon. Alternatively, one can obtain predictions for longer time horizons by training a temporal operator for large $\Delta t$, which can limit the accumulation of errors, but does not eliminate it entirely.

\section*{Temporal Neural Operator (TNO)} \label{TNO}

The Deep Operator Network (DeepONet)\cite{Lu2021LearningOperators} consists of two primary components: a branch network ($b_i$), which processes input functions, and a trunk network ($t_i$), which processes query locations. The learned operator \( \mathcal{G}_{\theta} \) approximates a nonlinear operator that maps an input function $v$ to its output at a query point $y$, and is expressed as:
\begin{equation}
    \mathcal{G}_{\theta}(v)(y) = \sum_{i=1}^{p} b_i \cdot t_i = \sum_{i=1}^{p} b_i(v(x_1), v(x_2), \ldots, v(x_m)) \cdot t_i(y).
    \label{eq:joined_outputDeepONet}
\end{equation}

The branch network takes as input a discretized representation of the function $v$, evaluated at a set of predefined  sensor points \( \{x_i\}_{i=1}^m \). This results in the input vector $[v(x_1), v(x_2), \ldots,$ $v(x_m)]^T$, which the branch network maps to a feature vector \( [b_1, b_2, \ldots, b_p]^T \in \mathbb{R}^p \). The trunk network takes the query location $y$ as input and outputs another feature vector \( [t_1, t_2, \ldots, t_p]^T \in \mathbb{R}^p \). Here, $p$ is the latent dimension of the learned representation space. The output of the DeepONet is the inner product of these two feature vectors which yields $\mathcal{G}_{\theta}(v)(y)$.

We extend the DeepONet framework by introducing a temporal branch (t-branch) that processes solution snapshots $u(t, \cdot)$.  This t-branch is designed to capture the temporal dynamics of the system and, when trained using temporal bundling, enables the model to effectively interpolate and extrapolate in time. The output of the proposed Temporal Neural Operator (TNO) is the predicted sequence of future states $\hat{U}_{fut}$, computed as a nonlinear function of the Hadamard product (element-wise multiplication) of the branch, trunk, and t-branch outputs. This combined representation is then projected into the output solution space of temporal length $K$. This architecture generalizes the DeepONet by incorporating temporal dynamics directly into the operator learning framework:

\begin{equation}
    \mathcal{G}_{\theta}^{L \rightarrow K}(\mathbf{U}_{\text{hist}}(t))(x) = G\left( b_i(v) \odot t_i(t, x) \odot tb_i(\mathbf{U}_{\text{hist}}(t)) \right), \quad \text{where } b_i, t_i, tb_i \in \mathbb{R}^p
    \label{eq:joined_outputTNO}
\end{equation}
where $G$ is the projection to the solution space of size $K$ in time. Here, $p$ denotes the latent dimension of the learned representation space, shared across the outputs of the branch, trunk, and temporal-branch networks. The Hadamard product combines these three $p$-dimensional feature vectors into a unified representation, which is then mapped by the projection operator $G : \mathbb{R}^p \rightarrow \mathbb{R}^{K \times d}$, parametrized as a multilayer perceptron (MLP), to predict the future solution sequence $\widehat{\mathbf{U}}_{\text{fut}}(t)$.

\begin{figure*}[!t]
    \centering
    \includegraphics[width=\textwidth]{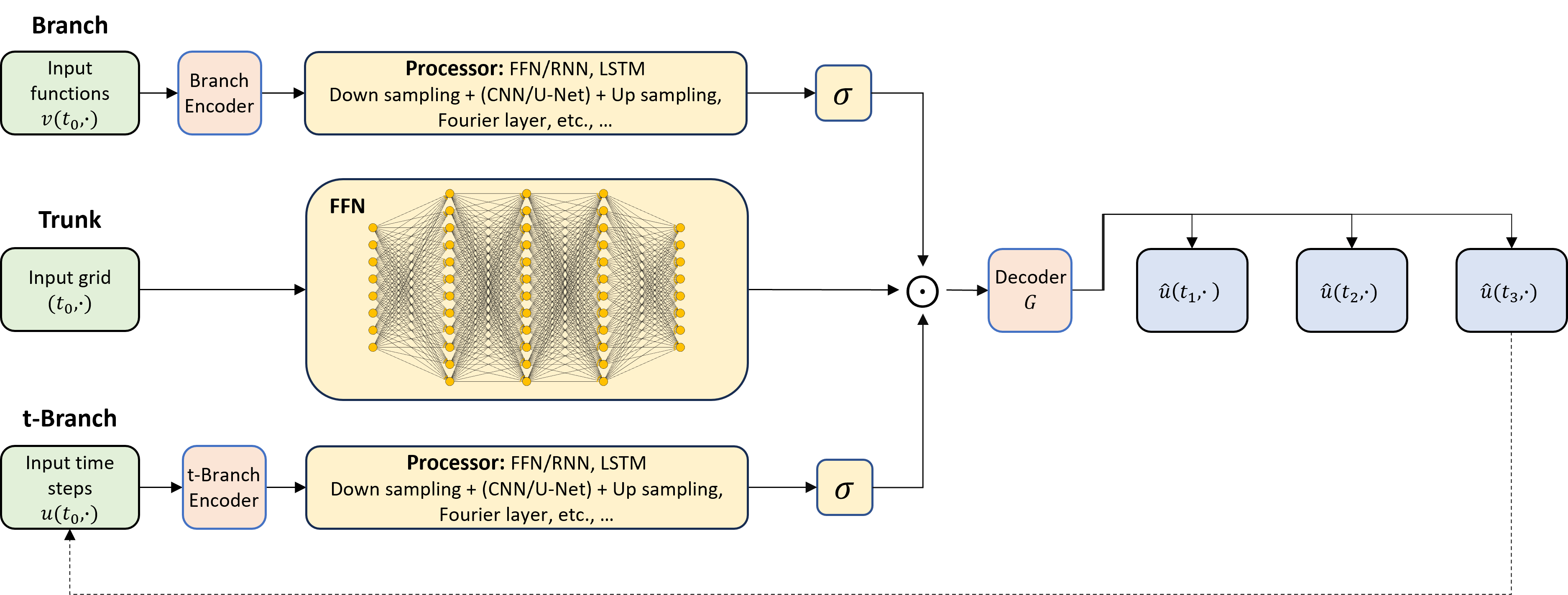}
    \caption{TNO architecture with temporal bundling training procedure. In this figure, \( L = 1 \) and \( K = 3 \)} as an example.
    \label{fig:TNOTrain}
    \vskip -0.2in
\end{figure*}

As an example with \( L = 1 \) and \( K = 3 \), the input to the TNO at time \( t_0 \) consists of the triplet \( \{ u(t_0, \cdot),\ (t_0, \cdot),\ v(t_0, \cdot) \} \), where \( (t_0, \cdot) \) represents the full spatio-temporal coordinate field, and \( v(t_0, \cdot) \) denotes an auxiliary time-dependent input function (e.g., a forcing term or coefficient field). The network predicts a bundle of future solution states \( \{ \hat{u}(t_1, \cdot),\ \hat{u}(t_2, \cdot),\ \hat{u}(t_3, \cdot) \} \); see Figure~\ref{fig:TNOTrain} for a visual summary of the training procedure using temporal bundling.

Temporal bundling enables the TNO to produce multiple future steps in a single forward pass, thereby improving efficiency and training stability. Once the initial bundle is predicted, the final output \( \hat{u}(t_3, \cdot) \) is reused as part of the next input for autoregressive rollout. During training, if teacher forcing is employed, the ground truth \( u(t_3, \cdot) \) is used instead to stabilize learning and mitigate error accumulation.

The architecture of the TNO is illustrated in Figure~\ref{fig:TNOTrain} and described in detail next. We remark here that we use a U-Net in both the branch and t-branch networks.

\begin{enumerate}
   
    \item Lift the input functions \( v(t_i, \cdot) \) and the full solution history \( \mathbf{U}_{\text{hist}}(t) = \{ u(t - (l-1)\Delta t, \mathbf{x}) \}_{l=1}^{L} \) to separate latent representation spaces using two independent linear encoders \( P_b \) and \( P_{tb} \):
    \begin{equation}
        \begin{aligned}
            h_b(v) &= P_b(v(t_i, \cdot)) \in \mathbb{R}^{p}, \\
            h_{tb}(\mathbf{U}_{\text{hist}}(t)) &= P_{tb}(\mathbf{U}_{\text{hist}}(t)) \in \mathbb{R}^{p}.
        \end{aligned}
        \label{eq:lift_to_latent_separate}
    \end{equation}
    The branch encoder processes the input function \( v \), while the t-branch encoder processes the full temporal history of the solution \( \mathbf{U}_{\text{hist}} \). Each encoder maps its input into a latent space of dimension \( p \), but these mappings are learned independently.

    \item Process the aligned latent representations through U-Net architectures in both the branch and t-branch. To handle variability in spatial resolution, the latent feature maps \( h_b(v) \) and \( h_{tb}(\mathbf{U}_{\text{hist}}(t)) \) are first passed through 2D adaptive average pooling to produce a fixed spatial resolution. The pooled features are then processed through separate U-Net blocks, followed by bilinear upsampling to restore the feature maps to the original spatial resolution \( H \times W \) of the input function or data grid:
    \begin{equation}
        \begin{aligned}
            q_b &= \text{AdaptiveAvgPool2d}(h_b(v)) \in \mathbb{R}^{p \times S \times S}, \\
            q_{tb} &= \text{AdaptiveAvgPool2d}(h_{tb}(\mathbf{U}_{\text{hist}}(t))) \in \mathbb{R}^{p \times S \times S}, \\
            U_b &= \text{U-Net}_b(q_b) \in \mathbb{R}^{p \times S \times S}, \\
            U_{tb} &= \text{U-Net}_{tb}(q_{tb}) \in \mathbb{R}^{p \times S \times S}, \\
            \tilde{U}_b &= \text{Upsample}(U_b) \in \mathbb{R}^{p \times H \times W}, \\
            \tilde{U}_{tb} &= \text{Upsample}(U_{tb}) \in \mathbb{R}^{p \times H \times W}.
        \end{aligned}
        \label{eq:unet_blocks}
    \end{equation}
    We found that it is often beneficial to set the pooling resolution \( S \times S \) to the lowest available input resolution, as this reduces memory usage and improves training efficiency. In practice, it may not be necessary to explicitly define or tune \( S \), as the resolution can be inferred from the data or heuristically selected.
    This combination of adaptive pooling, U-Net, and upsampling allows the network to generalize across inputs of varying spatial resolutions while maintaining high representational capacity. It decouples the input resolution from the network architecture, enabling the model to operate on arbitrary grid sizes during inference without retraining.

    \item Encode the spatio-temporal coordinates using the trunk network. The trunk takes as input the coordinates \( (t, \mathbf{x}) \), where \( t \in \mathbb{R} \) denotes the time, and \( \mathbf{x} \in \mathbb{R}^m \) denotes the spatial location. These coordinates are passed through a fully connected feedforward neural network \( f_\theta \), typically with nonlinear activation functions (e.g., hyperbolic tangent), to produce a feature vector in the latent space \( \mathbb{R}^p \):
    \begin{equation}
        t_i(\mathbf{x}, t) = f_\theta(\mathbf{x}, t) \in \mathbb{R}^{p \times H \times W}.
        \label{eq:trunk_network}
    \end{equation}
    The trunk features \( t_i(\mathbf{x}, t) \) serve as the coordinate-dependent embedding that interacts with the outputs of the branch and t-branch through an element-wise (Hadamard) product.

    \item The full TNO output can be written in operator form as:
    \begin{equation}
        \widehat{\mathbf{U}}_{\text{fut}}(t)(\mathbf{x}) = \mathcal{G}^{L \rightarrow K}_{\theta}(\mathbf{U}_{\text{hist}}(t))(\mathbf{x}) = G\left( \tilde{U}_b(\mathbf{x}, t) \odot \tilde{U}_{tb}(\mathbf{x}, t) \odot t_i(\mathbf{x}, t) \right),
        \label{eq:tno_operator_form}
    \end{equation}
    where \( \mathcal{G}^{L \rightarrow K}_{\theta} \) denotes the learned operator mapping a temporal history of \( L \) past solution states to a bundled prediction of \( K \) future time steps, \( \odot \) is the Hadamard product along the latent feature dimension \( p \), and \( G : \mathbb{R}^p \rightarrow \mathbb{R}^K \) is a shared MLP decoder applied pointwise over the spatial domain. The output \( \mathcal{G}^{L \rightarrow K}_{\theta}(\mathbf{U}_{\text{hist}}(t)) \in \mathbb{R}^{K \times H \times W} \) represents a temporally bundled solution over the full spatial grid.

\end{enumerate}
The TNO architecture is implemented in PyTorch and trained on an NVIDIA Tesla V100 GPU. The ADAM optimizer is used to minimize the mean square error (MSE) loss between the predicted and ground truth solutions. The same U-Net architecture is employed for both the branch and temporal branch across all examples. Additional U-Net architectural details are provided in Appendix~\ref{U-NEt_Arch}.

\section*{Results} \label{Results}

\subsection*{Weather Forecast for European Air Temperature} \label{WeatherForeCast}
The E-OBS (European Observations) dataset \cite{cornes2018ensemble} is a high-resolution observational climate dataset designed for analyzing climate variability and long-term trends across Europe. It covers a broad geographical region (approximately \( 25^\circ\text{N} - 71.5^\circ\text{N} \times 25^\circ\text{W} - 45^\circ\text{E} \)) and includes several decades of daily mean temperature measurements. A key feature of E-OBS is its high spatial resolution, available at 0.1° and 0.25°, corresponding to regular grids of size \( 705 \times 465 \) and \( 201 \times 464 \), respectively. This fine-grained spatial coverage enables detailed regional weather and climate analysis. Unlike datasets generated or constrained by numerical models, E-OBS is constructed directly from European weather station observations. As a result, the dataset presents unique challenges for learning-based models: spatial gaps, temporal discontinuities, and evolving coverage patterns introduce inconsistencies that the TNO must learn to handle during training and inference.

For this experiment, we train the TNO with an input bundle size of \( L = 1 \) and an output bundle size of \( K = 4 \), using daily mean temperature and pressure fields as input variables. The training set spans 9,000 days from 1997 to 2022. Validation is performed on 360 days from January 1, 2022, to December 26, 2022, and testing is conducted on 360 days from January 1, 2023, to December 26, 2023. All training and testing are conducted on the lower-resolution version of the dataset (0.25°, corresponding to a \( 201 \times 464 \) grid). To evaluate the TNO's resolution invariance, an additional high-resolution test set (0.1°, \( 705 \times 465 \)) is constructed for the period December 26, 2022, to December 31, 2023. This setup allows us to assess whether the TNO can generalize to unseen spatial resolutions without retraining.

The training, validation, and testing datasets are standardized using z-score normalization computed from per-pixel statistics of the training set. Each training sequence is structured into bundles of nine time steps: the first snapshot serves as the initial condition, followed by two rollout windows of size four for autoregressive supervision. More details on the training procedure and hyperparameters are provided in Appendix \ref{Tr_Hyp_European}. This example highlights several key capabilities of the TNO architecture:

\begin{itemize}
    \item \textbf{Forecasting under real-world observational noise and gaps:} The TNO effectively learns from gridded observational data with missing values and temporal-spatial inconsistencies, showcasing robustness to incomplete or imperfect datasets.
    
    \item \textbf{Multi-day sequence prediction with minimal historical context:} With an input sequence length of \( L = 1 \), the TNO successfully generates forecasts over a multi-day horizon (\( K = 4 \)), demonstrating its ability to model short-term atmospheric dynamics with limited history.

    \item \textbf{Generalization to unseen high-resolution spatial grids:} The TNO accurately forecasts temperature fields on a higher-resolution grid (0.1°) at test time, despite being trained exclusively on lower-resolution data (0.25°), indicating resolution-invariant generalization.
\end{itemize}

\begin{figure}[t]
    \centering
    % Subfigure 1
    \begin{subfigure}[b]{1\linewidth}
        \centering
        \includegraphics[width=\linewidth]{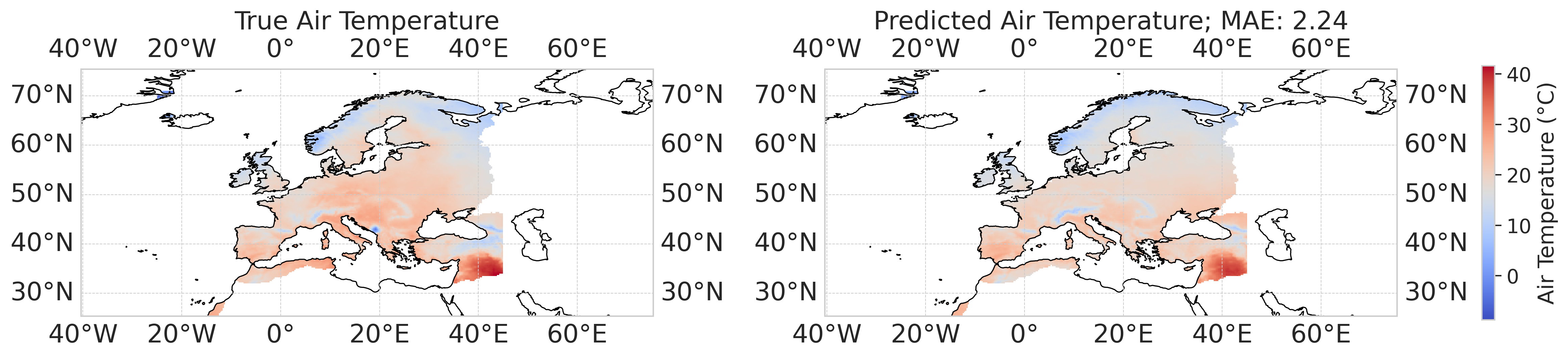}
        \caption{Air temperature (24/06/2023, 0.25° grid)}
        \label{fig:air_temp_155}
    \end{subfigure}

    % Subfigure 1
    \begin{subfigure}[b]{1\linewidth}
        \centering
        \includegraphics[width=\linewidth]{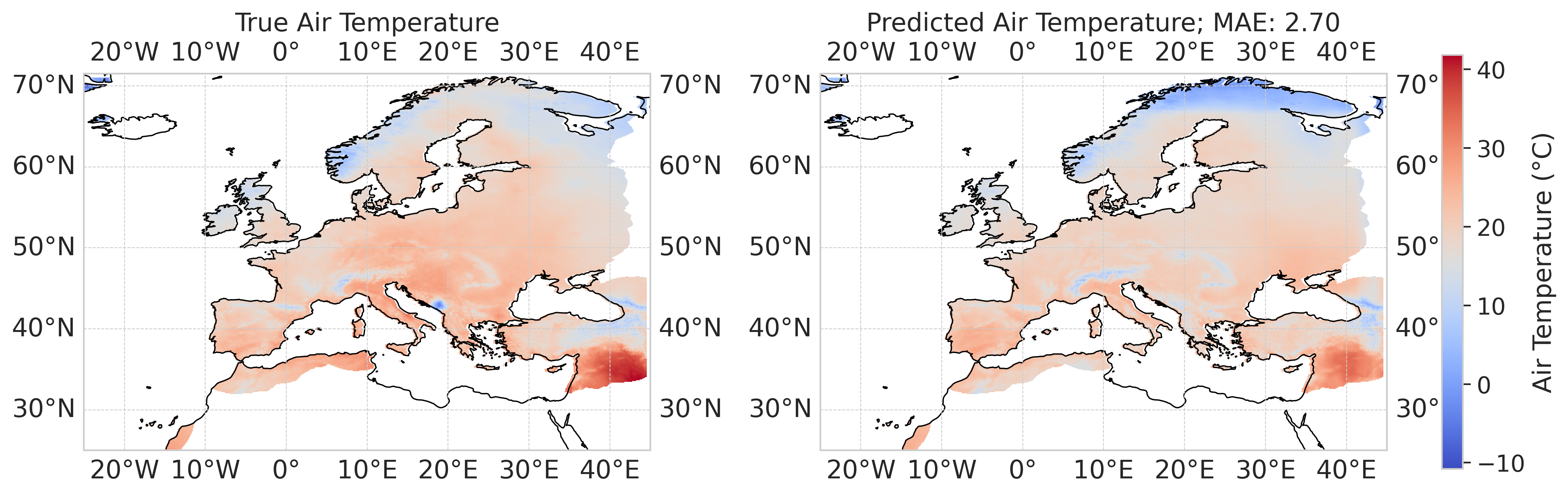}
        \caption{Air temperature (24/06/2023, 0.1° grid).}
        \label{fig:fine_air_temp_155}
    \end{subfigure}

    \caption{Qualitative Testing performance of the TNO - Summer 2023}
    \label{fig:test_temp_june}
\end{figure}

\begin{figure}[t]
    \centering
    % Subfigure 2
    \begin{subfigure}[b]{1\linewidth}
        \centering
        \includegraphics[width=\linewidth]{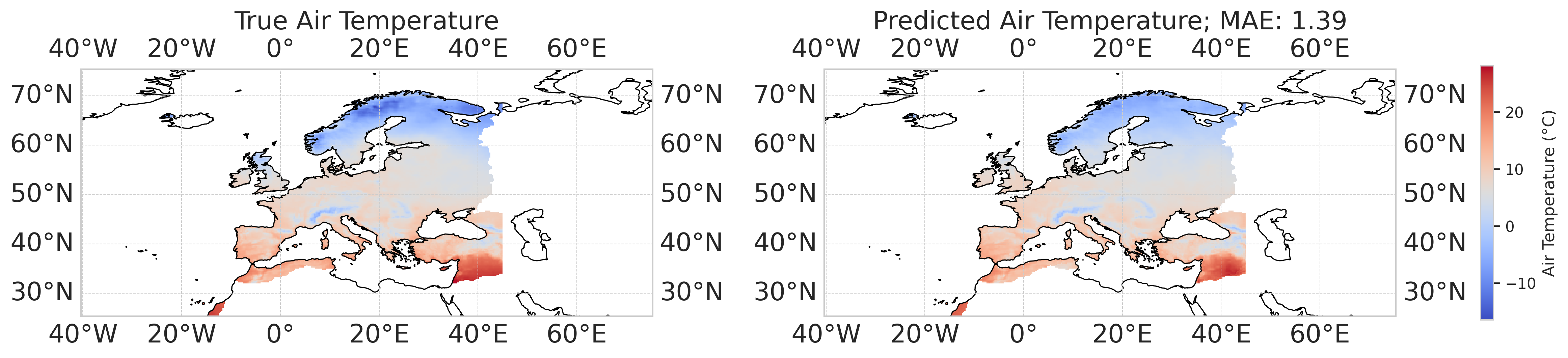}
        \caption{Air temperature (12/11/2023, 0.25° grid)}
        \label{fig:air_temp_280}
    \end{subfigure}
    
    % Subfigure 2
    \begin{subfigure}[b]{1\linewidth}
        \centering
        \includegraphics[width=\linewidth]{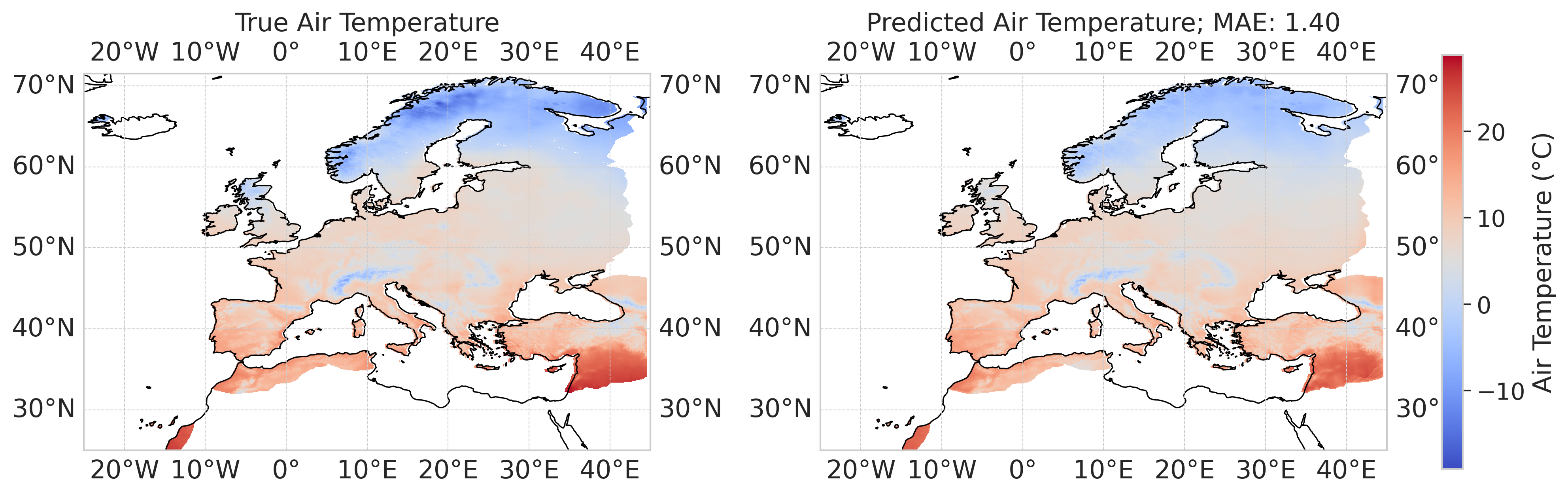}
        \caption{Air temperature (12/11/2023, 0.1° grid).}
        \label{fig:fine_air_temp_280}
    \end{subfigure}

    \caption{Qualitative Testing performance of the TNO - Winter 2023}
    \label{fig:test_temp_nov}
\end{figure}

\subsubsection*{Results of Weather Forecast for European Air Temperature} \label{ResultsWeatherForeCast}
To evaluate the TNO performance for regional weather forecasting, we utilize a blind test dataset from the year 2023. During testing, the TNO is initialized with initial conditions comprising daily mean air temperature, pressure, and the spatial grid. The model then performs two autoregressive rollouts of size \( 2 \times K \), predicting air temperatures across successive eight-day intervals. After each prediction cycle, a new initial condition is drawn from the ground truth to initialize the next rollout window. This approach simulates a realistic deployment setting, where updated observational data is periodically supplied and the model forecasts the near-future temperature evolution.

The mean absolute error (MAE) and root mean square error (RMSE) of the predicted temperatures per time snapshot is shown in \ref{fig:Benchmarks}, with an overall MAE of \( 2.68^\circ \mathrm{C} \) across all snapshots. The results indicate that the TNO does not incur significant error accumulation over successive rollouts. Two qualitative examples of predicted air temperatures for May 06, 2023 and November 12, 2023 are shown in Figures~\ref{fig:test_temp_june}\subref{fig:air_temp_155} and~\ref{fig:test_temp_nov}\subref{fig:air_temp_280}, respectively, both of which demonstrate excellent agreement with the ground truth. These findings underscore the TNO’s effectiveness in temporal extrapolation and its ability to maintain stable prediction quality over extended forecasting horizons.

To assess the TNO’s resolution invariance, we employ a second testing dataset for the same period (2023), rendered at a higher resolution of 0.1°. Remarkably, the TNO achieves an overall MAE of \( 2.83^\circ \mathrm{C} \) on this dataset—without any additional training or fine-tuning—highlighting the model’s capacity to generalize across spatial resolutions. The per-snapshot MAE and RMSE for the high-resolution test set is shown in Figure \ref{fig:Benchmarks}, and corresponding predicted temperature fields for May 06, 2023 and November 12, 2023 are visualized in Figures~\ref{fig:test_temp_june}\subref{fig:fine_air_temp_155} and~\ref{fig:test_temp_nov}\subref{fig:fine_air_temp_280}, respectively. These results further demonstrate the robustness and versatility of the TNO in handling variations in spatial discretization while maintaining high predictive accuracy. Additional results for this dataset are provided in Appendix~\ref{Add_Rslts_European}.

\subsubsection*{Ablation Study} \label{sec:ablation}
To asses the contribution of the individual components in our TNO, we evaluate three ablated variants alongside the full TNO on the 2023 coarse‑grid test set (0.25°):

\begin{itemize}
  \item \textbf{TNO without t‑Branch}: the temporal branch is removed, so predictions rely only on the branch (U‑Net) and trunk.
  \item \textbf{TNO without U‑Net}: both the branch and temporal branch U‑Net blocks are replaced by MLPs.
  \item \textbf{One step TNO without t‑Branch}: combines the previous two ablations (no t‑branch and \(K=1\)).
  \item \textbf{One step TNO}: the full TNO architecture (t‑branch, U‑Net) but predicting one step at a time (\(K=1\)), i.e.\ no temporal bundling.
  \item \textbf{Full TNO}: the complete architecture as presented, with t‑branch, U‑Net spatial encoders, and temporal bundling (\(L=1,K=4\)).
\end{itemize}

Figure~\ref{fig:Benchmarks} presents per‑snapshot RMSE and MAE for these variants on both the coarse (top row) and fine (bottom row) grids. On the coarse grid, removing the temporal branch (\texttt{TNO without tBranch}) roughly doubles both RMSE and MAE and also eliminates the model’s grid‑invariance capability. Replacing the U‑Net with MLPs (\texttt{TNO without UNet}) leads to a significant increase in spatial error, underscoring the importance of deep spatial feature extraction. The combined ablation (\texttt{OneStep TNO without tBranch}) performs comparably to \texttt{TNO without tBranch}, indicating that temporal bundling alone cannot compensate for the absence of the t‑branch. Overall, the Full TNO achieves the lowest RMSE and MAE across all forecast snapshots, validating the complementary roles of the U‑Net, temporal branch, and temporal bundling.

\subsubsection*{Benchmark Comparisons} \label{sec:benchmarks_comp}
We compare the full TNO against several state‑of‑the‑art neural operators on the 2023 test sets at both coarse (0.25°) and fine (0.1°) resolutions. The baselines are:

\begin{itemize}
  \item \textbf{DeepONet (one‑step):} Vanilla DeepONet trained autoregressively to predict one day ahead (\(K=1\)), without temporal bundling.
  \item \textbf{DeepONet (multi‑step):} Vanilla DeepONet with temporal bundling (\(L=1,K=4\)), predicting four days in each forward pass.
  \item \textbf{Fourier‑DeepONet} \cite{zhu2023fourier}: A Fourier Neural Operator variant adapted for temporal extrapolation with the same bundling (\(L=1,K=4\)).
  \item \textbf{TNO (full):} The Temporal Neural Operator with t‑branch, U‑Net spatial encoders, and temporal bundling (\(L=1,K=4\)).
\end{itemize}

\begin{figure}[hp]
    \centering
    \includegraphics[width=0.8\linewidth]{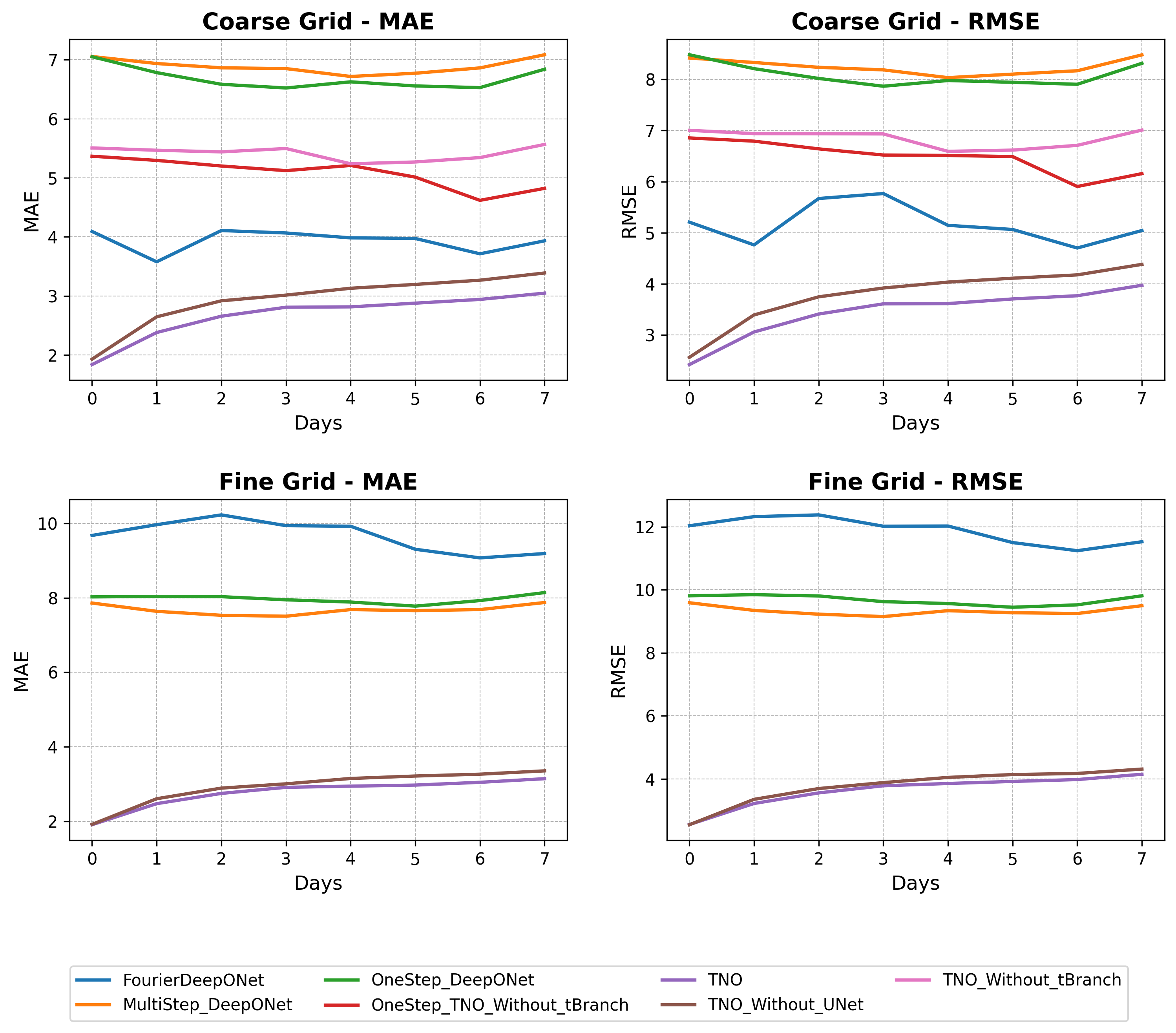}
    \caption{MAE (left) and RMSE (right) on the coarse (top) and fine (bottom) grids. “Multistep” denotes temporal bundling (\(K=4\)), “Onestep” predicts one step at a time.}
    \label{fig:Benchmarks}
\end{figure}

Figure~\ref{fig:Benchmarks} (top row) reports per‐snapshot RMSE and MAE for these methods on the coarse grid, while the bottom row reports similar results for the fine grid predictions. On the coarse grid, the Full TNO consistently achieves the lowest MAE and RMSE at all lead times, outperforming both OneStep and Multi step DeepONets and the Fourier-DeepONet by a wide margin. When evaluated at the higher 0.1° resolution without any retraining, the Full TNO’s error increases only marginally, whereas Fourier-DeepONet and the DeepONet variants suffer substantial degradation. These results demonstrate the TNO’s superior accuracy, stability over long horizons, and robustness to changes in spatial resolution.

\subsection*{Climate Modeling for Global Air Temperature} \label{GAT}

Modeling the Earth's climate involves simulating complex, nonlinear, and multi-scale dynamics across interacting systems such as the atmosphere, oceans, land, and ice. Traditional approaches, like General Circulation Models (GCMs) \cite{GCM1991}, solve time-dependent partial differential equations (PDEs) governing fluid flow, heat transfer, and radiation. They are computationally expensive and rely on parameterizations for unresolved processes fraught with uncertainty. Recent machine learning advancements, including foundation models \cite{bodnar2024aurora, pathak2022fourcastnet}, offer data-driven alternatives but still require vast datasets and compute budgets \cite{chen2023foundation}. In contrast, the TNO presented here is designed as a lightweight, computationally efficient framework for solving time-dependent PDEs directly from spatio-temporal data, making it a practical tool for scientific modeling in resource-constrained settings.

Our objective in this experiment is to leverage the TNO to efficiently and accurately model the spatio-temporal evolution of global air temperature, while reducing computational costs and enabling forecasts across all atmospheric pressure levels. This experiment demonstrates three core capabilities of the TNO:

\begin{itemize}
    \item \textbf{Long-term temporal extrapolation:} The TNO accurately forecasts global air temperature fields over a five-year horizon, demonstrating strong extrapolation performance beyond the temporal range seen during training.
    
    \item \textbf{Vertical generalization across pressure levels:} The model successfully interpolates and extrapolates temperature fields at atmospheric pressure levels not included in the training set, highlighting its ability to generalize across the vertical dimension.
    
    \item \textbf{Computationally efficient 3D modeling via 2D operations:} By treating the 3D spatio-temporal climate data as a collection of 2D slices conditioned on pressure levels, the TNO leverages 2D convolutional architectures to efficiently learn in a high-dimensional setting.
    
    \item \textbf{Unified multi-level forecasting:} Unlike models trained on individual pressure levels, the TNO uses a single architecture to forecast temperature fields across all levels simultaneously, this reduces model complexity and improves generalization.

\end{itemize}

\begin{figure*}[htbp]
    \centering
    % Subfigure 1
    \begin{subfigure}[b]{0.9\linewidth}
        \centering
        \includegraphics[width=\linewidth]{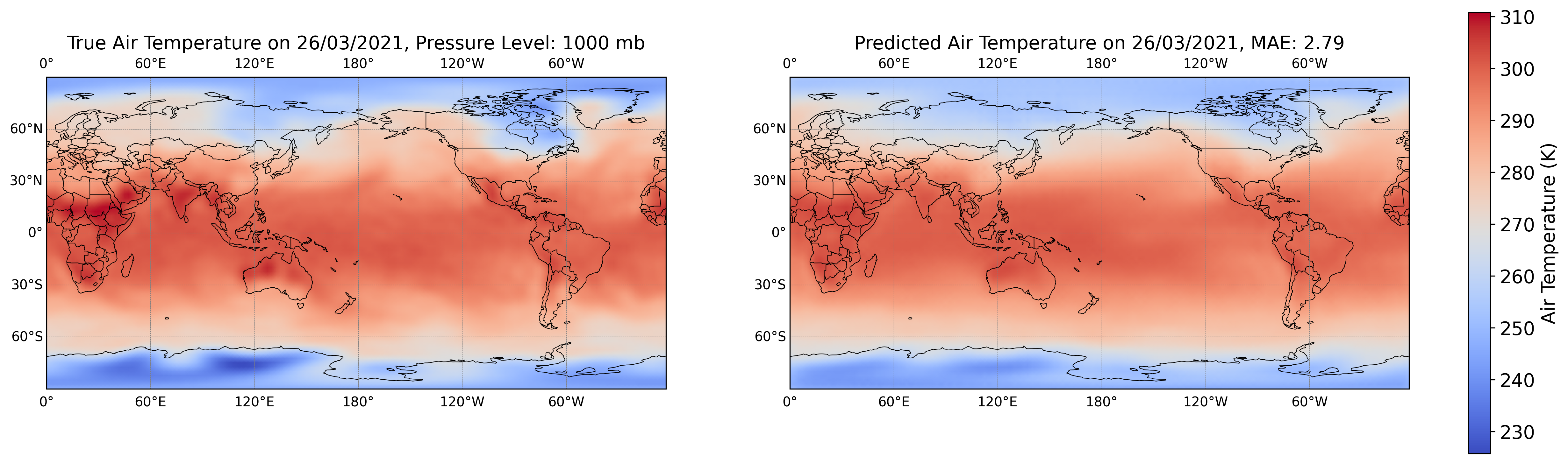}
        \vskip -0.1in
        \caption{Global air temperature comparison (26/03/2021, at 1000 mb pressure).}
        \label{fig:26_03_2021_Pressure_1000}
    \end{subfigure}

    % Subfigure 2
        \begin{subfigure}[b]{0.9\linewidth}
        \centering
        \includegraphics[width=\linewidth]{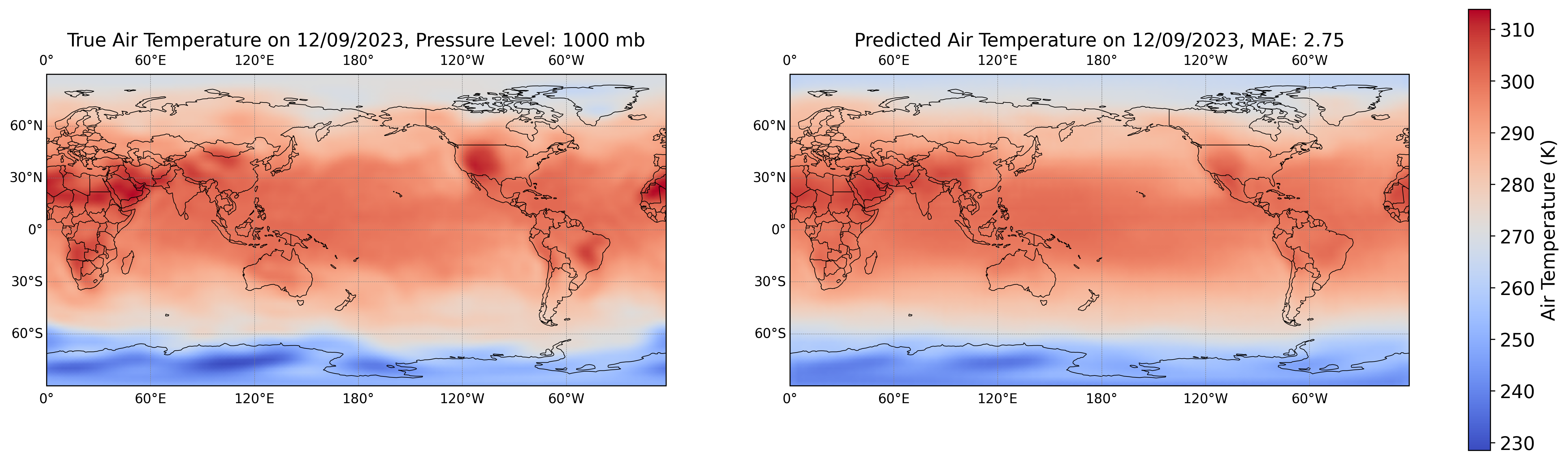}
        \vskip -0.1in
        \caption{Global air temperature comparison (12/09/2023, at 1000 mb pressure).}
        \label{fig:12_9_2023_Pressure_1000}
    \end{subfigure}
    
    % Subfigure 3
    \begin{subfigure}[b]{0.9\linewidth}
        \centering
        \includegraphics[width=\linewidth]{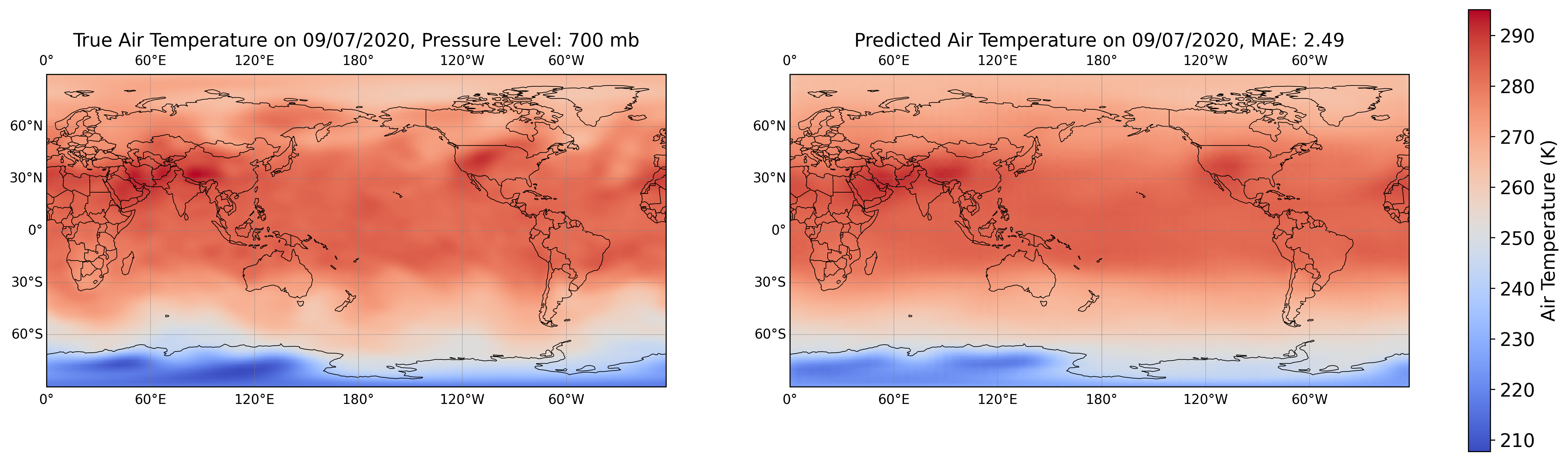}
        \vskip -0.1in
        \caption{Global air temperature comparison (9/07/2020, at 700 mb pressure).}
        \label{fig:9_7_2020_Pressure_700}
    \end{subfigure}

    % Subfigure 3
    \begin{subfigure}[b]{0.9\linewidth}
        \centering
        \includegraphics[width=\linewidth]{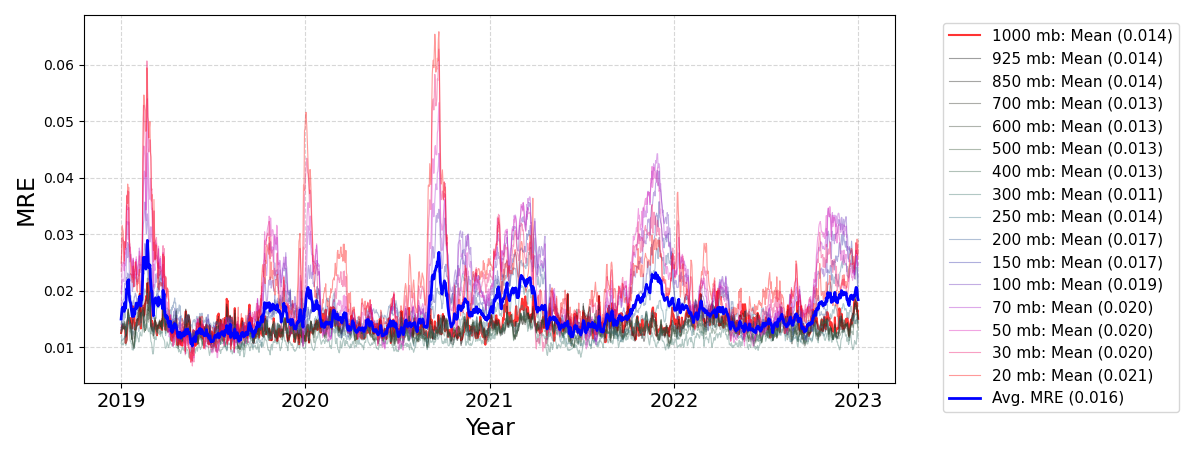}
        \vskip -0.1in
        \caption{Mean Relative $L_2$ Error (MRE) per time snapshot of the testing dataset on every pressure level starting from January 1, 2020 to December 31, 2023. The blue line represents the average MRE for all pressure levels.}
        \label{fig:MRE_Climate}
    \end{subfigure}
    
    \caption{Testing performance of the TNO for Global Temperature Forecast. Note that the initial condition of year 2019 are excluded from the error calculation.}
    \label{fig:Climate_Pressure1000}
    \vskip -0.2in
\end{figure*}

To evaluate these capabilities, we use the NCEP/NCAR Reanalysis 1 dataset \cite{kalnay2018ncep}, which provides daily mean air temperature fields on a global \( 144 \times 72 \) spatial grid (2.5° resolution) across 16 pressure levels, spanning from 1948 to the present. For this experiment, we use data from January 1, 2010 to December 31, 2015 for training, January 1, 2016 to December 31, 2018 for validation, and January 1, 2019 to December 31, 2023 for testing.

The temporal branch receives a history of air temperature fields \( \mathbf{T}_{\text{hist}}(t) \), while the branch network is conditioned on the atmospheric pressure level \( \mathbf{P} \). This formulation allows the TNO to both interpolate and extrapolate air temperature across pressure levels, while treating the 3D temperature field as a series of 2D slices. To demonstrate this, we train the TNO using data from 12 out of the 16 available pressure levels: $ \{1000,\ 925,\ 850,\ 600,\ 300,\ 250,\ 200,\ 150,\ 70,\ 50,\ 30,\ 20\} \ \text{mb}, $ holding out 4 pressure levels for evaluation.

The dataset was standardized using z-score normalization computed from per-pixel statistics of the training set. We define the temperature field at time \( t \) and pressure level index \( p \) as \( \mathbf{T}_p(x, y, t) \), where \( (x, y) \) denotes spatial coordinates on the global grid. For each training example, we construct an input tensor \( \mathbf{T}_{\text{hist}, p} \in \mathbb{R}^{H \times W \times L} \), representing the temperature field at pressure level \( p \) over \( L \) past time steps. The corresponding target is a tensor \( \hat{\mathbf{T}}_{\text{fut}, p} \in \mathbb{R}^{H \times W \times K} \), representing predicted temperatures at the same level over the next \( K \) time steps. In this experiment, we set both the input and output sequence lengths to one year: \( L = K = 365 \). During training, the TNO is provided with five years of daily temperature data, spanning from January 1, 2010 to December 31, 2015. The model is trained across 12 distinct pressure levels from the available dataset, with each level treated as a separate input condition to the branch network. To improve training efficiency and promote cross-level generalization, pressure levels are batched in groups of four. That is, during each training step, the TNO receives a batch of four temperature tensors \( \mathbf{T}_{\text{hist}, p} \) corresponding to four distinct pressure levels, and jointly predicts the corresponding future sequences \( \hat{\mathbf{T}}_{\text{fut}, p} \). For each year and pressure level \( p \), the temporal branch receives an input tensor \( \mathbf{T}_{\text{hist}, p} \in \mathbb{R}^{144 \times 72 \times 365} \) from year \( t \), and the TNO predicts \( \hat{\mathbf{T}}_{\text{fut}, p} \) for year \( t + 1 \), using the corresponding ground truth as supervision. This process is repeated sequentially across all training years and level batches. Finally, we use teacher forcing during training, where ground truth sequences are used to guide future prediction and accelerate convergence. During inference, the TNO generates multi-step forecasts by recursively using its own predictions as input. Additional training details and hyperparameter configurations are provided in Appendix \ref{Tr_Hyp_GlobalAir}.

\subsubsection*{Results of Global Air Temperature Forecast} \label{ResultsGlobalForeCast}
To evaluate the performance of the TNO, we use a blind test set comprising five years of daily mean air temperature data, spanning all 16 atmospheric pressure levels
($ \{1000,\ 925,\ 850,\ 700,\ 600,\ 500,\ 400,\ 300,\ 250,\ 200,\ 150,\ 100,\ 70,\ 50,\ 30,\ 20\} \ \text{mb}, $)
from January 1, 2019 to December 31, 2023. This setup is designed to test the TNO's ability to (i) extrapolate temperature fields forward in time, and (ii) interpolate and extrapolate across pressure levels not included in training. The evaluation metric used is the relative \( L_2 \) error, computed across all spatial grid points and time steps.

This task represents a challenging generalization scenario in which the TNO must forecast into an unseen temporal domain and make predictions at pressure levels it has not encountered during training. As shown in Figure~\ref{fig:Climate_Pressure1000}\subref{fig:MRE_Climate}, the TNO achieves strong performance in both respects, with a mean relative \( L_2 \) error of 0.016. The error is generally higher at higher-altitude (lower-pressure) levels, likely due to data imbalance across pressure levels in the training set.

Additionally, at 1000 mb—the lowest atmospheric level—the TNO achieves accuracy on par with the Ditto transformer \cite{ovadia2023ditto}, which was trained specifically for that level and for the same extrapolation period. However, unlike Ditto, the TNO generalizes across all levels with a single unified model. Figure~\ref{fig:Climate_Pressure1000}\subref{fig:26_03_2021_Pressure_1000} shows a visual comparison between the TNO's predictions and the ground truth air temperature fields at 1000 mb for selected dates from the testing period (2020–2023). Additional qualitative and quantitative results are provided in Appendix \ref{Add_Rslts_GlobalAir}.

\begin{figure}[t]
\centering
\begin{subfigure}[b]{0.45\textwidth}
  \centering
  \includegraphics[width=1\linewidth]{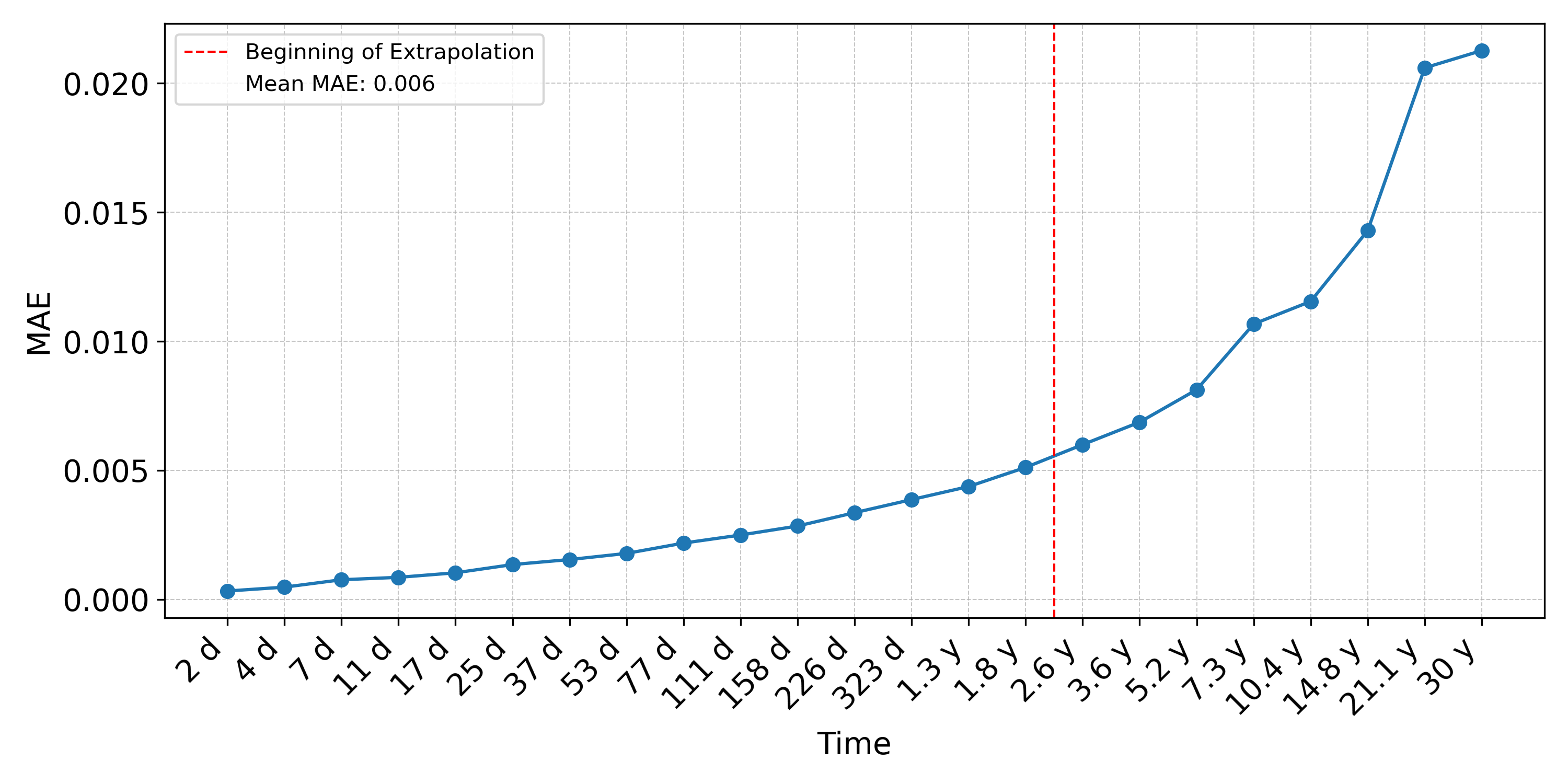}
  \caption{$\mathrm{CO}_2$ Saturation plume area MAE}
  \label{fig:sat}
\end{subfigure}
\hfill % Use "\hfill" for maximum spacing between the two figures
\begin{subfigure}[b]{0.45\textwidth}
  \centering
  \includegraphics[width=1\linewidth]{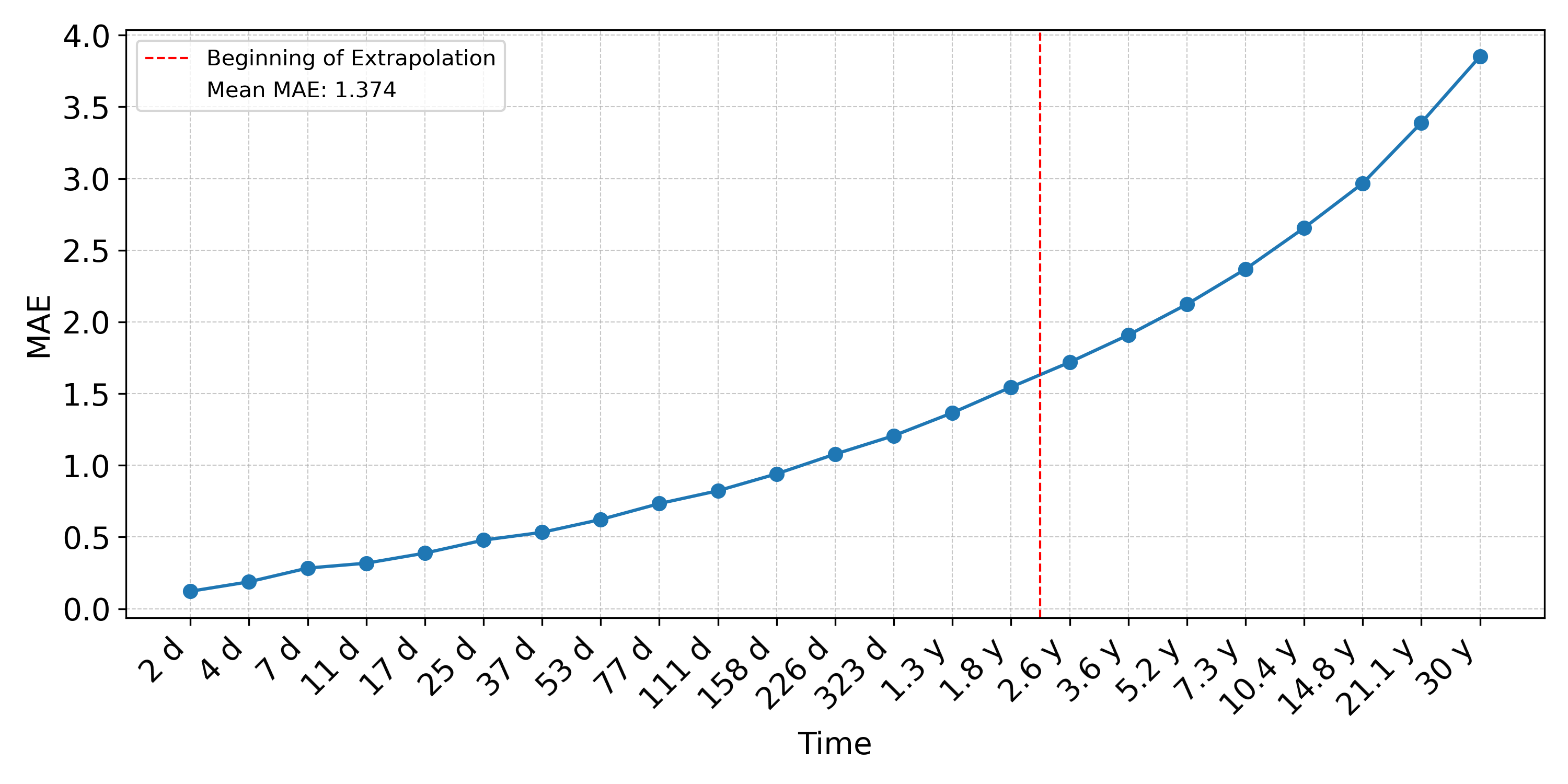}
  \caption{$\mathrm{CO}_2$ Pressure Buildup MAE}
  \label{fig:dP}
\end{subfigure}

\caption{Evaluation metrics of simultaneous generalization and extrapolation to unseen realizations and time snapshots for all $\mathrm{CO}_2$ saturation and pressure buildup (dP) time snapshots. Extrapolation begins after 1.8 years. Results are averaged over 500 realizations.}
\label{fig:MAE_GS&dP_Extrap}
\end{figure}

\subsection*{Geologic Carbon Sequestration} \label{CarbonSeq}

In geological carbon sequestration (GCS), captured CO\(_2\) is injected into deep subsurface formations for long‑term storage, thereby reducing atmospheric emissions \cite{EnergyAgency2023NetUpdate}. Accurate predictions of CO\(_2\) plume migration and pressure evolution are essential for ensuring containment, optimizing injection strategies, and satisfying regulatory monitoring requirements. Modeling GCS processes involve complex, coupled multiphase flow and transport governing PDEs. Simulations are computationally demanding, as they must span decades and account for site‑specific heterogeneities  \cite{Wang2023AnReservoirs,Pruess2002MultiphaseAquifers, Mahjour2023SelectingUncertainty}.

In this final example, we employ the TNO to model $\mathrm{CO}_2$ plume migration and pressure buildup in an underground geological storage site. We utilize the dataset published in \cite{Wen2022U-FNOAnFlow} to evaluate the TNO's performance for GCS problems. The same dataset was used in our earlier work \cite{Diab2023U-DeepONet:Sequestration} to train various neural operator architectures; however, none of those architectures, including \cite{Wen2022U-FNOAnFlow}, were able to extrapolate well beyond the temporal training horizon of the dataset. In this work, we evaluate the TNO’s ability to (i) extrapolate plume migration and pressure buildup over extended timescales, and (ii) generalize across varying geological conditions, embedding both the elliptic and hyperbolic PDE dynamics in its learned operator.

\subsubsection*{Training and Inference for GCS} \label{Training_GCS}
We base our experiments on the benchmark GCS dataset from \cite{Wen2022U-FNOAnFlow}, which comprises 5,500 realizations of input–output mappings for gas saturation \(S_g\) and pressure buildup \(\Delta P\). Each realization contains solutions at 24 successively coarsening time snapshots 
$\{1\,\text{day},2\,\text{days},4\,\dots,323\,\text{days},1.3\,\text{years},\dots,21.1\,\text{years},30\,\text{years}\}$.
The original split (9:1:1) allocates 4,500 realizations for training, 500 for validation, and 500 for testing. To evaluate long‑term extrapolation, we restrict both training and validation to the first 16 snapshots $\{1\ \text{day},\ldots,1.8\ \text{years}\}$, reserving the final 8 snapshots $\{2.6\ \text{years},\ldots,30\ \text{years}\} $ for blind testing of temporal extrapolation and generalization.

Each input realization consists of four spatial fields—horizontal permeability \(\mathbf{k}_x\), vertical permeability \(\mathbf{k}_y\), porosity \(\boldsymbol{\phi}\), and perforation height \(\mathbf{h}_{\mathrm{perf}}\) (each \(\in\mathbb{R}^{H\times W}\))—and five scalars: initial reservoir pressure \(P_{\text{init}}\), injection rate \(Q\), temperature \(T\), capillary scaling \(\lambda\), and irreducible water saturation \(S_{wi}\). The outputs are the saturation field \(\mathbf{S}_g\in\mathbb{R}^{H\times W}\) and pressure buildup field \(\Delta\mathbf{P}\in\mathbb{R}^{H\times W}\). We direct the readers to the original paper \cite{Wen2022U-FNOAnFlow} for more details on the reservoir description, the numerical simulation, the generation of the field maps and all other sampling techniques for the inputs.

Our objective in this experiment is to leverage the TNO to efficiently and accurately model the spatio‑temporal evolution of CO\(_2\) plume migration and pressure buildup in subsurface formations, while reducing computational cost and enabling long‑term forecasts under heterogeneous geological conditions. This experiment demonstrates three core capabilities of the TNO:
\begin{itemize}
  \item \textbf{Long‑term temporal extrapolation:} Evaluate forecasting accuracy for saturation and pressure fields over multi‑decadal horizons beyond the training time window (\(t > 1.8\) years).
  \item \textbf{Generalization to unseen geological conditions:} Test robustness to new input functions including diverse permeability, porosity fields, and well‑configuration not seen during training.
  \item \textbf{Coupled multiphysics modeling:} Examine the TNO’s capacity to concurrently capture elliptic pressure dynamics and hyperbolic saturation transport within a unified operator framework.
\end{itemize}

Following the TNO architecture (Fig.~\ref{fig:TNOTrain}), the four input fields and five scalars are encoded by the branch network; the initial condition \( \mathbf{S}_g(x,y,t_0),\,\Delta\mathbf{P}(x,y,t_0)\) is processed by the t‑branch, and the temporal grid is handled by the trunk. We set the history length and forecast horizon to \(L=1\) and \(K=3\), respectively, during training and testing.

\subsubsection*{Results of Geologic Carbon Sequestration} \label{ResultsGCS}
\begin{figure}[htbp]
    \centering
    \includegraphics[width=0.8\linewidth]{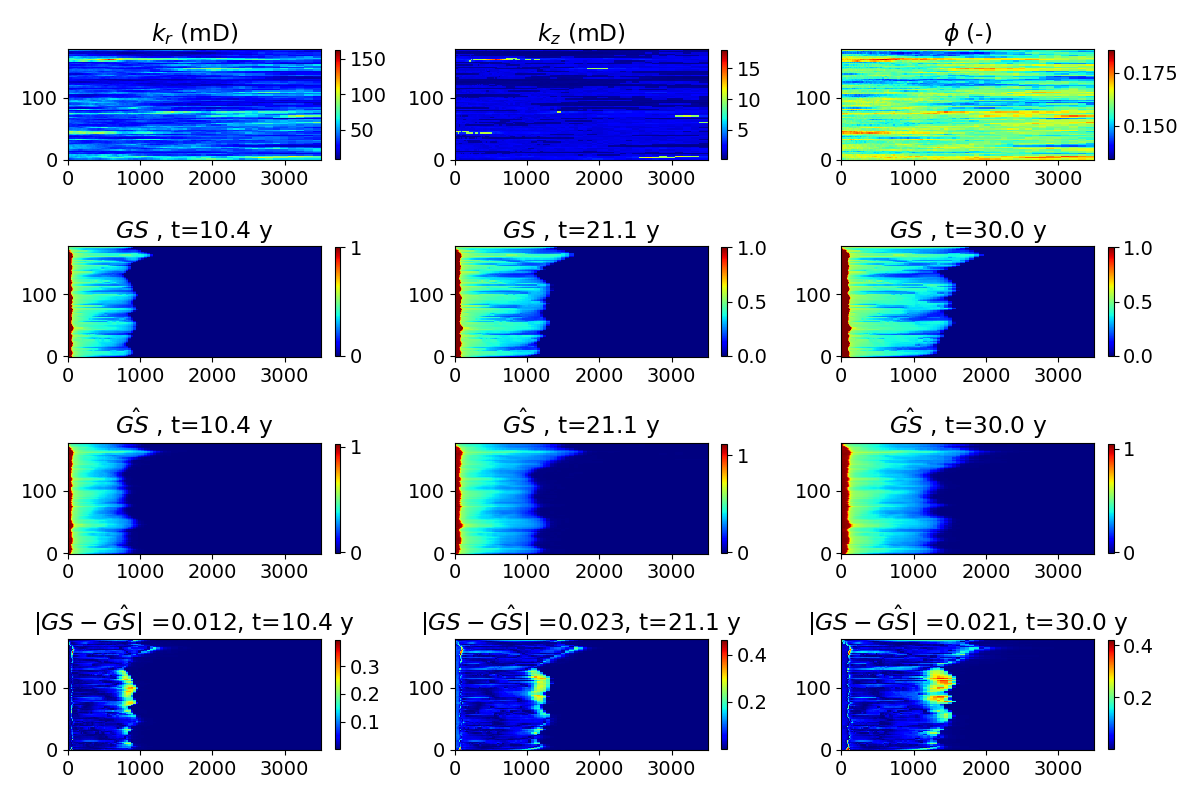}
    \caption{An example of $\mathrm{CO}_2$ saturation performance in simultaneous extrapolation and generalization. \textit{Row 1}: Input reservoir properties: radial permeability (\(k_r\)), vertical permeability (\(k_z\)), and porosity (\(\phi\)).  
    \textit{Row 2}: Ground truth saturation field.  
    \textit{Row 3}: TNO predicted saturation.  
    \textit{Row 4}: Absolute error.}
    \label{fig:LongTerm_Extrap_Test_GS}
\end{figure}

\begin{figure}[htbp]
    \centering
    \includegraphics[width=0.8\linewidth]{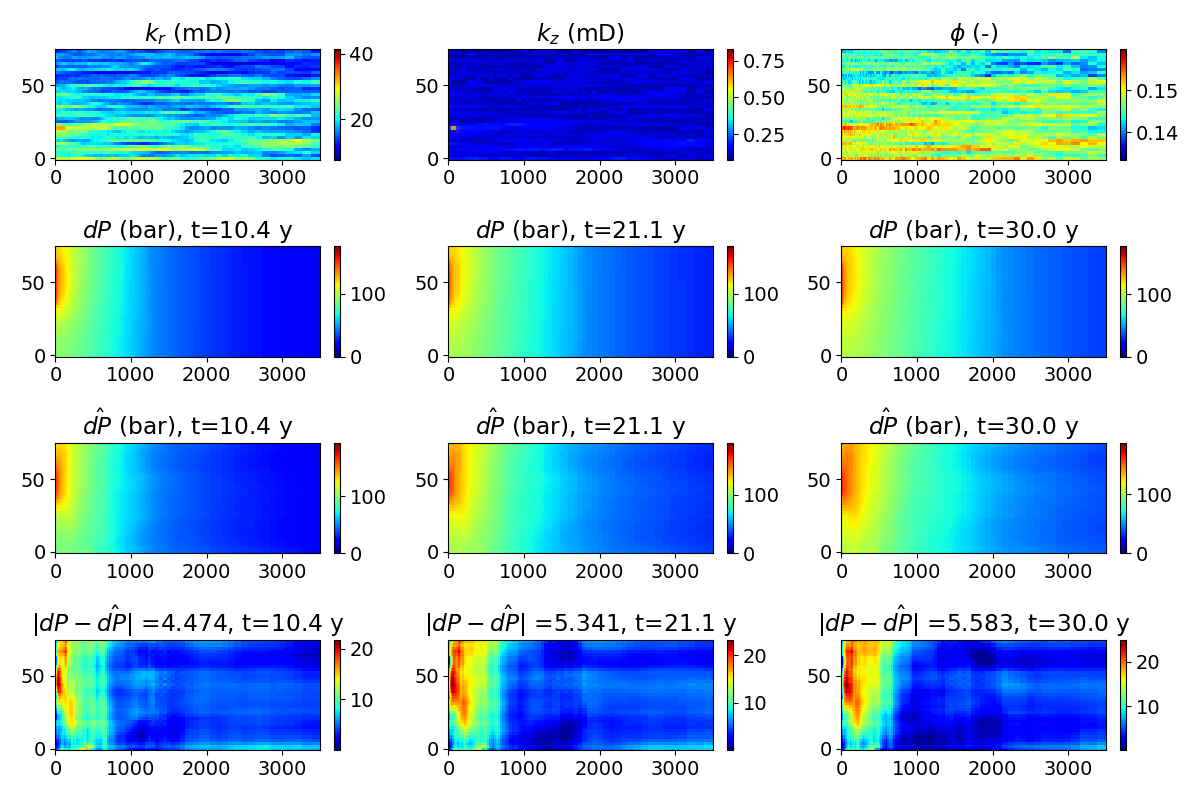}
    \caption{An example of pressure buildup (dP) performance in simultaneous extrapolation and generalization. \textit{Row 1}: Input reservoir properties: radial permeability (\(k_r\)), vertical permeability (\(k_z\)), and porosity (\(\phi\)).  
    \textit{Row 2}: Ground truth pressure buildup field.  
    \textit{Row 3}: TNO predicted pressure buildup.  
    \textit{Row 4}: Absolute error.}
    \label{fig:Lonfterm_Extrap_Test_dP}
\end{figure}

To assess the performance of the TNO in simultaneous generalization and temporal extrapolation tasks for both $\mathrm{CO}_2$ saturation and pressure buildup, we utilize the testing dataset which consists of 500 previously unseen realizations. Importantly, time snapshots beyond 1.8 years were excluded during training, which allows us to classify performance on any time snapshot within the training time range as generalization, and performance beyond 1.8 years as simultaneous generalization and temporal extrapolation. The mean absolute error serves as the primary metric for quantifying TNO performance. For $\mathrm{CO}_2$ saturation specifically, error calculations are restricted to the $\mathrm{CO}_2$ saturation plume region, as saturation values outside this area remain zero. Figure \ref{fig:MAE_GS&dP_Extrap} shows the MAE for the saturation \ref{fig:MAE_GS&dP_Extrap}\subref{fig:sat} and for the pressure buildup \ref{fig:MAE_GS&dP_Extrap}\subref{fig:dP}. Both figures clearly demonstrate that the TNO effectively generalizes and extrapolates to unseen realizations of gas saturation and pressure buildup, as well as to time snapshots not included in the training data. There is, however, a slight decline in solution quality towards the end of the temporal domain, which is to be expected since we are further out in the extrapolation. Figures \ref{fig:LongTerm_Extrap_Test_GS} and \ref{fig:Lonfterm_Extrap_Test_dP} show an example of $\mathrm{CO}_2$ saturation plume and $\mathrm{CO}_2$ pressure buildup towards the end of the extrapolation period, respectively. Additional results for $\mathrm{CO}_2$ saturation plume and pressure buildup can be found in Appendices \ref{Results_GCS_GS} and \ref{Results_GCS_dP}, respectively.

\section*{Discussion} \label{Discussion}
This paper introduces the Temporal Neural Operator (TNO) for reliable generalization and temporal extrapolation of parametric time-dependent PDEs. The TNO's performance was evaluated across three distinct problem domains: forecasting global air temperature across all atmospheric levels for climate modeling, predicting regional air temperatures for Europe and adjacent regions, and modeling flow and transport in geologic carbon sequestration, involving the tracking of $\mathrm{CO}_2$ saturation plume and pressure buildup. Despite the diverse nature of these problems and the unique challenges posed by their respective datasets, the TNO consistently demonstrated remarkable performance. These results further highlight our model's versatility in handling complex spatio-temporal dynamics across various scientific applications.

The TNO addresses several longstanding challenges in Scientific Machine Learning (SciML). Namely, simultaneous long-term extrapolation and generalization, grid invariance, computational efficiency and low memory foot-print. The TNO is a lightweight, efficient architecture that can operate seamlessly on a single off-the-shelf GPU. For example, in the climate modeling problem, the TNO utilizes only 7.5 GB of GPU memory (see Table \ref{tab:performance_summary} for a complete summary on TNO performance). The TNO requires about 3.63 seconds per epoch (batch size of 4) for a total training time of only 7.2 minutes. Moreover, we take advantage of the generalization properties of the TNO to interpolate between atmospheric levels, which reduces the complexity of the 3D problem to a 2D learnable problem without any loss of information. By feeding the spatio-temporal grid to the trunk of the TNO, we learn the temporal dimension through a feed-forward neural network, further reducing the dimensionality of the problem. Importantly, the dimensionality reduction techniques applied in this example are confined to the learning process and do not alter the problem formulation itself. This level of flexibility and adaptability underscores the TNO's robust learning capacity and efficiency.

\begin{table}[h]
    \caption{Performance Summary of the TNO.}
    \label{tab:performance_summary}
    \centering
    \begin{tabular}{l|c|c|c}
        \hline
        \textbf{Dataset} & \textbf{Trainable Parameters (Million)}  & \textbf{GPU Memory (GB)} & \textbf{Time / Epoch (s)} \\
        \hline
        Global Temperature Forecast       & 0.12  & 6.4    & 39.0 \\
        European Temperature Forecast     & 39.4  & 12.2   & 3.63 \\
        Carbon Sequestration - GS         & 2.66  & 4.3    & 211.0 \\
        Carbon Sequestration - dP         & 7.41  & 7.5    & 447.0 \\
        \hline
    \end{tabular}
\end{table}

In the weather forecasting problem, the TNO grid invariance capabilities were demonstrated. The TNO is trained on air temperature data of resolution 0.25° taken from the E-OBS (European Observations) dataset \cite{cornes2018ensemble}. TNO is then tested on unseen data with a resolution of 0.1°. The accuracy in the temporal extrapolation task is quite remarkable given that no additional training or fine-tuning was needed, despite the inherent challenges in the dataset due to missing sensor data in some daily measurements and the changing areal coverage of the dataset.    

In the geological carbon sequestration problem, the TNO was trained on data representing the first two years of $\mathrm{CO}_2$ storage, corresponding to 16 time steps. Despite the limited temporal scope of the training data and the inherent complexity of the problem, the TNO demonstrated exceptional capabilities. It effectively generalized to new instances of the problem, which involved nine varying PDE parameters, while simultaneously extrapolating temporal dynamics for up to 30 years. This performance highlights the TNO's robustness and reliability, even in scenarios with limited data.

\bibliography{sample}

\section*{Acknowledgments}

The authors wish to acknowledge Khalifa University's high-performance computing facilities for providing computational resources.

\section*{Author contributions statement}

Waleed Diab: Conceptualization, Methodology, Software, Formal analysis, Investigation, Validation, Visualization, Writing – original draft, Writing – review and editing. Mohammed Al Kobaisi: Supervision, Conceptualization, Methodology, Formal analysis, Validation, Writing – original draft, Writing – review and editing. 

\section*{Competing interests}
The authors declare that they have no known competing financial interests or personal relationships that could have appeared to influence the work reported in this paper.

\section*{Accession codes}

The code will be made available upon request. All data used in this study is publicly available and cited appropriately where applicable.

\newpage
\appendix

\onecolumn
\section{U-Net Architecture} \label{U-NEt_Arch}
The U-Net architecture in our TNO follows an encoder-decoder structure with 2D convolutional and transposed convolutional layers for feature extraction and reconstruction. Activations used throughout the network are Leaky ReLU $(\alpha = 0.1)$ or Sigmoid Linear Unit (SiLU). The choice of activation function is problem specific and has a small, but measurable, effect on accuracy. The encoder consists of stride-2 convolutions with batch normalization, while the decoder employs stride-2 transposed convolutions with skip connections to preserve spatial information. The final output layer applies a 1×1 convolution. The network is initialized using Xavier initialization \cite{glorot2010understanding} for improved training stability. Table \ref{tab:unet_summary} provides a detailed summary of the network’s structure. 

\begin{table}[h]
    \caption{Summary of the U-Net Architecture}
    \label{tab:unet_summary}
    \vskip 0.15in
    \centering
    \begin{tabular}{c|c|c|c|c|c}
        \hline
        \textbf{Operation} & \textbf{Kernel Size} & \textbf{Stride} & \textbf{Activation} & \textbf{Normalization} & \textbf{Skip Connection} \\
        \hline
        Conv2D  & 3×3  & 2  & Activation  & Batch Norm  & No  \\
        Conv2D  & 3×3  & 2  & Activation  & Batch Norm  & Yes \\
        Conv2D  & 3×3  & 2  & Activation  & Batch Norm  & Yes \\
        Conv2D  & 3×3  & 1  & Activation  & Batch Norm  & No  \\
        Deconv2D  & 4×4  & 2  & Activation  & No  & Yes \\
        Deconv2D  & 4×4  & 2  & Activation  & No  & Yes \\
        Deconv2D  & 4×4  & 2  & Activation  & No  & Yes \\
        Conv2D  & 3×3  & 1  & None  & No  & No  \\
        \hline
    \end{tabular}
    \vskip -0.1in
\end{table}

\section{Training and Hyperparameters of the European Air Temperature Dataset} \label{Tr_Hyp_European}
For our first example, we utilize the E-OBS (European Observations) dataset \cite{cornes2018ensemble} which is a high-resolution dataset depicting climate variability and long-term trends across Europe. It covers a broad geographic area (roughly: $25N-71.5N \times 25W-45E$), spanning several decades, and includes daily mean temperature measurements collected directly from a network of European weather stations.
To train the TNO, the dataset is refactored into batches of 9 days, yielding a total of 1080 trajectories (batches). This results in tensor dimensions of (1000, 201, 464, 9) for training, (40, 201, 464, 9) for validation, and (40, 201, 464, 9) for testing.

Training the TNO for daily temperature forecast is carried out in two phases: \textit{teacher forcing} and \textit{fine-tuning}. In the teacher forcing phase, $L$ is chosen form the ground-truth data, while for the fine-tuning phase, $L = 1$ and chosen from the model predictions. In the teacher forcing phase, each prediction bundle of size $K = 4$ is conditioned on the ground truth data of the preceding rolled-out window. This allows the model to learn from the true data distribution without compounding errors. In the fine-tuning phase, the TNO learns to generate predictions relying on its own outputs as inputs. This allows the model to freely roll-out without any ground truth data. A training batch consists of four pieces of information: the spatio-temporal grid, initial pressure, initial temperature, and the subsequent 8 temperatures readings. Furthermore, a masking array is used to ignore invalid or missing regions in the ground truth data during training. The $L_{2}$-loss is computed over the unmasked region. The optimizer we use is Adam, which is initialized with a learning rate of $1 \times 10^{-3}$ and weight decay of $1 \times 10^{-3}$, with a cosine annealing learning rate schedule. The teacher forcing phase consists of 60 epochs, followed by 40 epochs of fine-tuning. The average training time per epoch is approximately 39 seconds. Moreover, each encoder projects its input into a 20-dimensional feature vector (i.e., we set \(p = 20\)), where \(p\) denotes the dimension of the linear encoder outputs in equation \ref{eq:lift_to_latent_separate}. Both U-Nets of the branch and the t-branch utilize the Sigmoid Linear Unit (SiLU) activation function, while the trunk utilizes the hyperbolic tangent (tanh) activation function.

\newpage
\section{Additional Results for the European Air Temperature Dataset} \label{Add_Rslts_European}

\begin{figure}[!htb]
    \centering
    % Subfigure 1
    \begin{subfigure}[b]{1\linewidth}
        \centering
        \includegraphics[width=\linewidth]{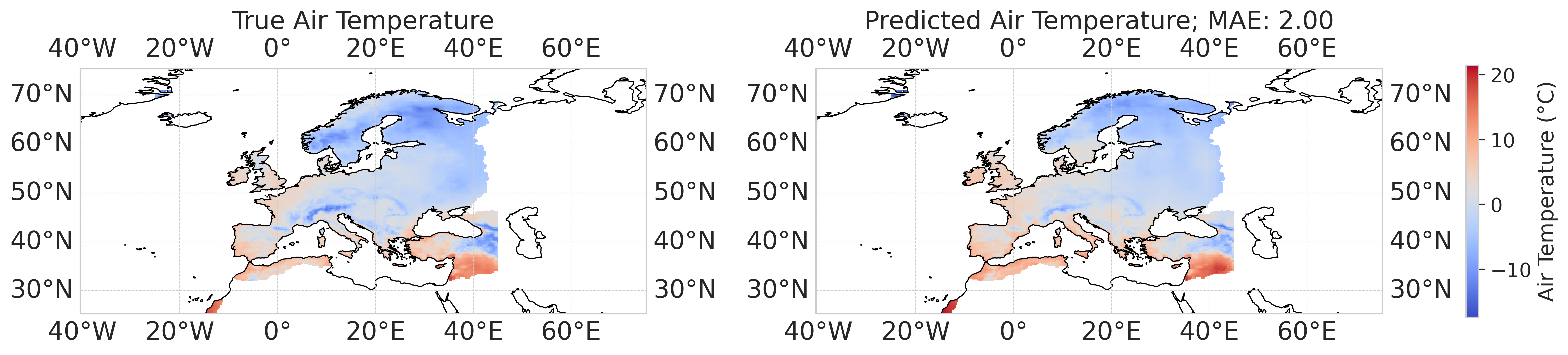}
        \caption{Air temperature (29/01/2023, 0.25° grid).}
        \label{fig:fine_air_temp_25_low}
    \end{subfigure}

    % Subfigure 2
    \begin{subfigure}[b]{1\linewidth}
        \centering
        \includegraphics[width=\linewidth]{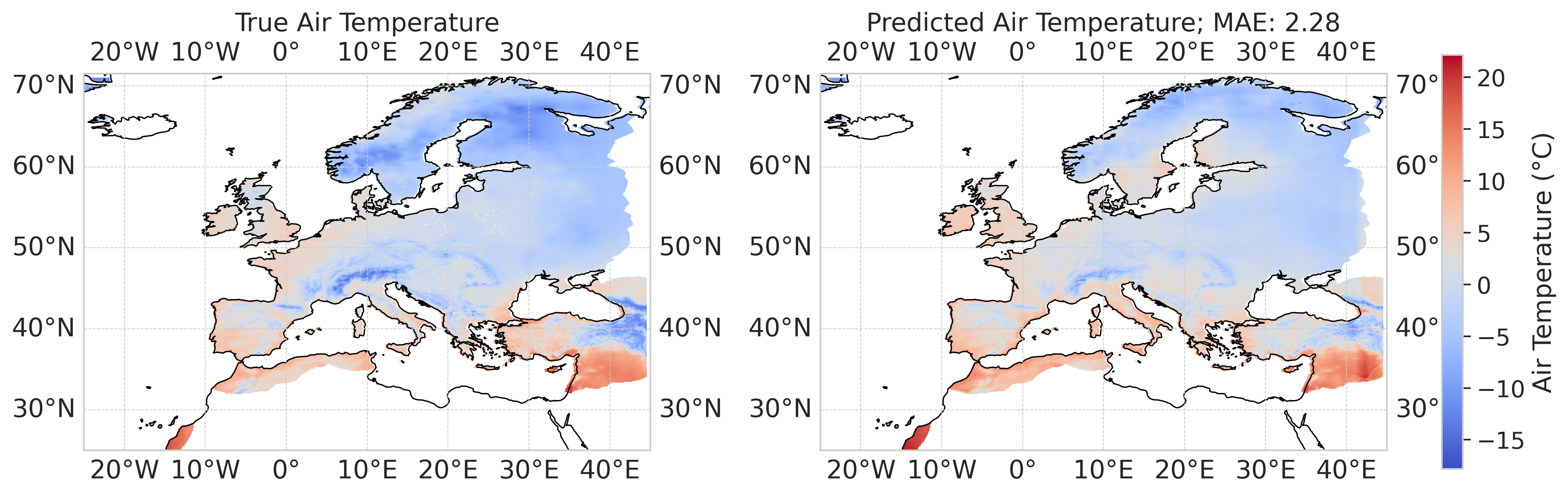}
        \caption{Air temperature (29/01/2023, 0.1° grid).}
        \label{fig:fine_air_temp_25_high}
    \end{subfigure}

    \caption{Testing performance of the TNO on 29/01/2023 at 0.25° and the corresponding solution at 0.1° resolutions.}
    \label{fig:fine_test_temp_25}
\end{figure}

\begin{figure}[!htb]
    \centering
    % Subfigure 1
    \begin{subfigure}[b]{1\linewidth}
        \centering
        \includegraphics[width=\linewidth]{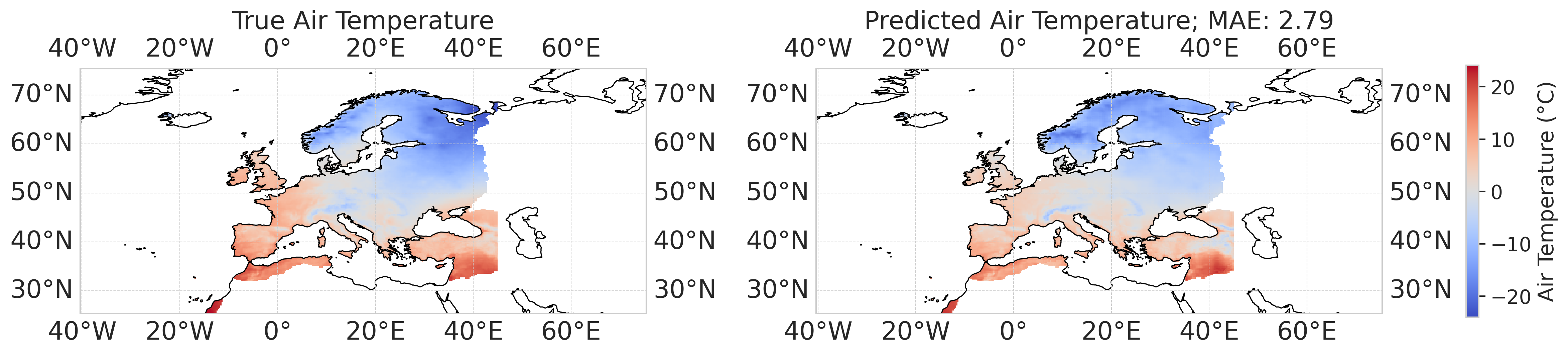}
        \caption{Air temperature (10/12/2023, 0.25° grid).}
        \label{fig:fine_air_temp_305_low}
    \end{subfigure}

    % Subfigure 2
    \begin{subfigure}[b]{1\linewidth}
        \centering
        \includegraphics[width=\linewidth]{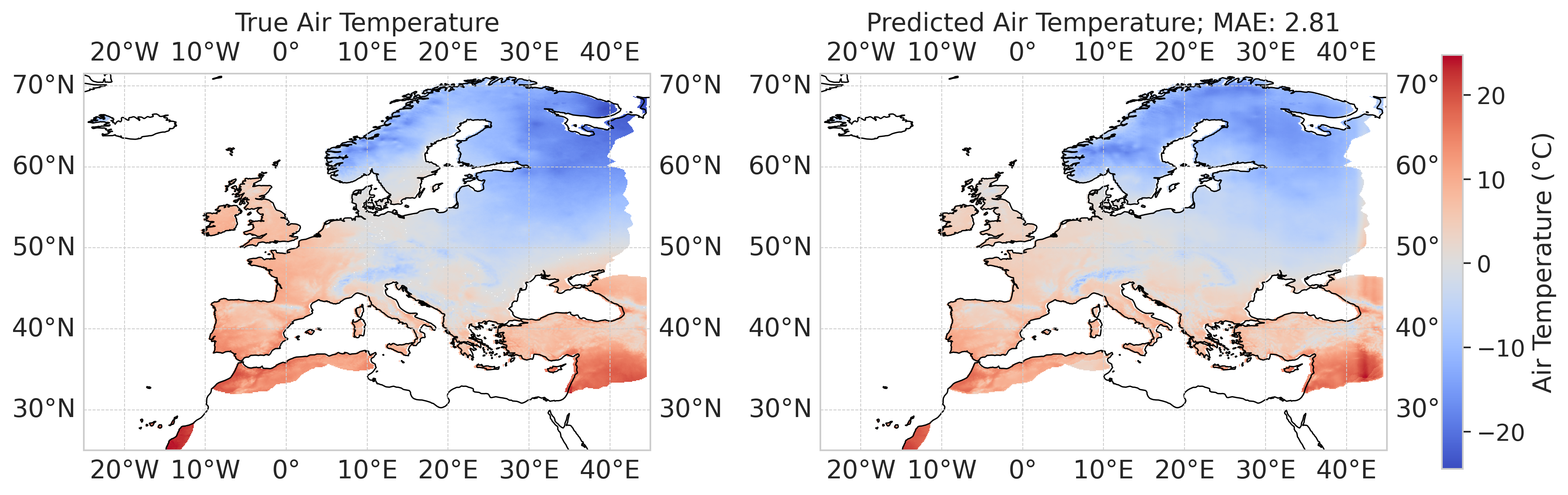}
        \caption{Air temperature (10/12/2023, 0.1° grid).}
        \label{fig:fine_air_temp_305_high}
    \end{subfigure}

    \caption{Testing performance of the TNO on 29/01/2023 at 0.25° and the corresponding solution at 0.1° resolutions.}
    \label{fig:fine_test_temp_305}
\end{figure}

\newpage
\section{Training and Hyperparameters of the Global Air Temperature Dataset} \label{Tr_Hyp_GlobalAir}

The NCEP/NCAR Reanalysis 1 dataset \cite{kalnay2018ncep} was used in our second example and it provides global daily mean air temperature values on a 2.5° resolution spatial grid. The dimensionality of the data during training is (Levels, Height, Width, Time), where we batch the data across the levels dimension, and we choose a batch size of 4. The TNO model is trained for 120 epochs;  air temperatures at various atmospheric levels are passed to the temporal-branch, the computational grid is passed to the trunk, and the atmospheric pressure levels are passed to the branch. At each iteration, the loss is minimized using the Adam optimizer with a learning rate of $1 \times 10^{-3}$ and a weight decay of $1 \times 10^{-4}$. A \textit{StepLR} scheduler reduces the learning rate by $10 \%$ every 5 epochs. The training process iteratively predicts air temperature fields for the predefined future forecast length $K$. For each segment, the model generates predictions that are compared to ground truth data over the next time window. Moreover, each encoder projects its input into a 360-dimensional feature vector (i.e., we set \(p = 360\)), where \(p\) denotes the dimension of the linear encoder outputs in equation \ref{eq:lift_to_latent_separate}. Both U-Nets of the branch and the t-branch utilize Sigmoid Linear Unit (SiLU) activation functions, while the trunk utilizes the hyperbolic tangent (tanh) activation function. The output of the branch is passed through a hyperbolic tangent (tanh) activation, and the output of the t-branch is passed through a Rectified Linear Unit (ReLU) activation.

\newpage
\section{Additional Results for the Global Air Temperature Dataset} \label{Add_Rslts_GlobalAir}

\begin{figure}[htbp]
    \centering
    % Subfigure 1
    \begin{subfigure}[b]{1\linewidth}
        \centering
        \includegraphics[width=\linewidth]{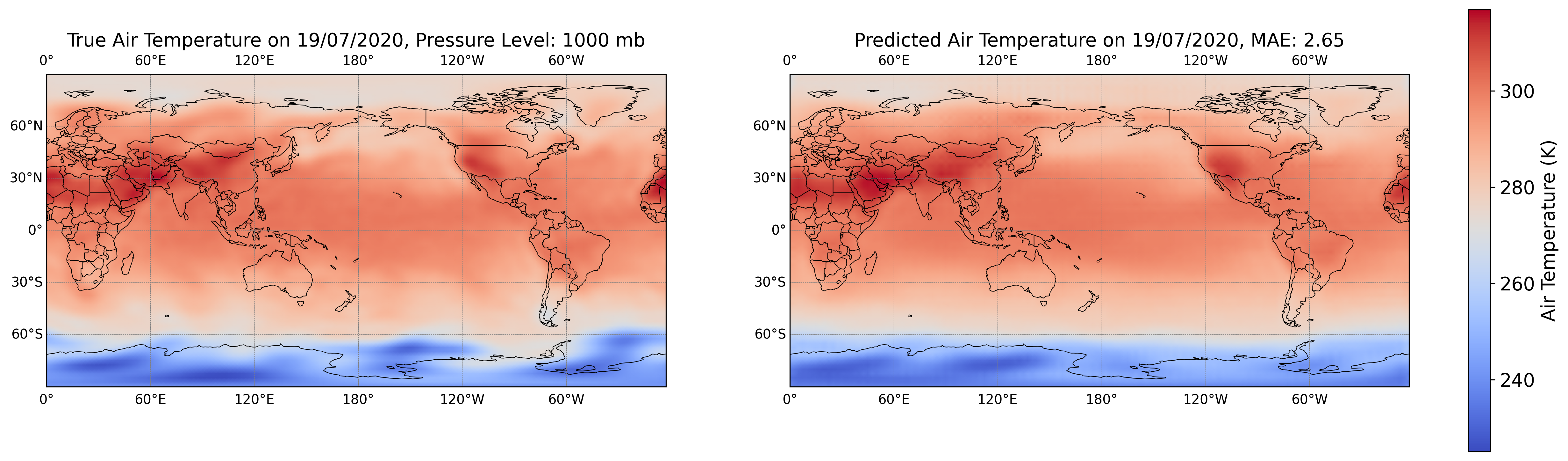}
        \caption{Global air temperature on 19/07/2020 and 1000 mb pressure.}
        \label{fig:19_7_2020_Pressure_1000}
    \end{subfigure}
    \vspace{-2pt} % Adjusts spacing between subfigures
    
    % Subfigure 3
    \begin{subfigure}[b]{1\linewidth}
        \centering
        \includegraphics[width=\linewidth]{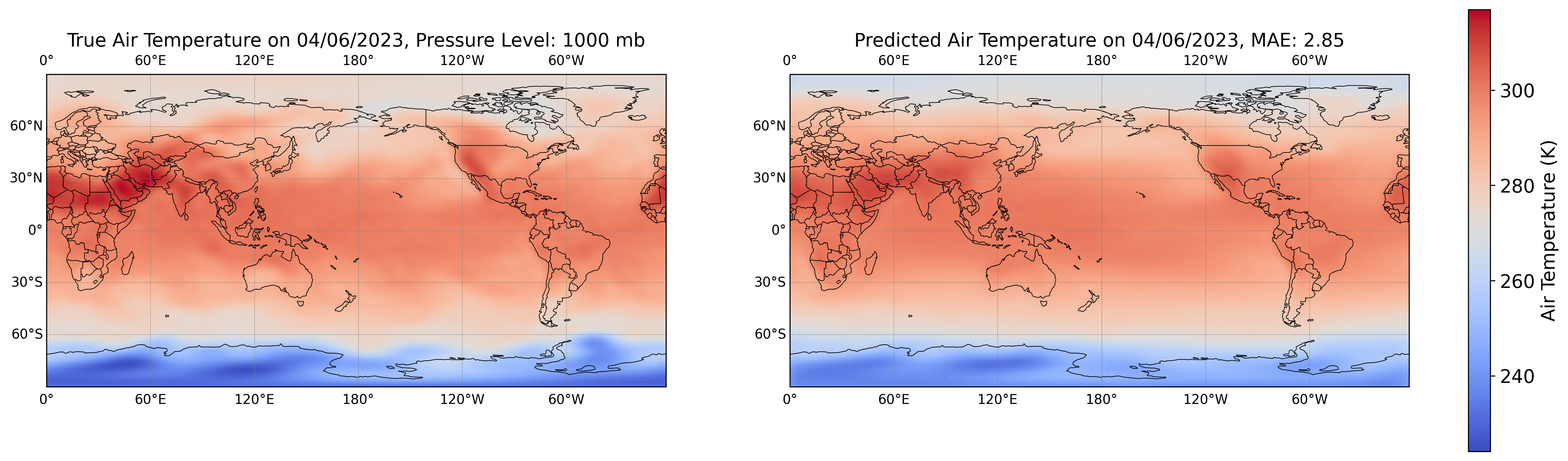}
        \caption{Global air temperature on 04/06/2023 and 1000 mb pressure.}
        \label{fig:4_6_2023_Pressure_1000}
    \end{subfigure}
    
    \caption{Testing performance of the TNO for Global Temperature Forecast.}
    \label{fig:Climate_Pressure1000_Apdx}
\end{figure}

\begin{figure}[htbp]
    \centering
    % Subfigure 1
    \begin{subfigure}[b]{1\linewidth}
        \centering
        \includegraphics[width=\linewidth]{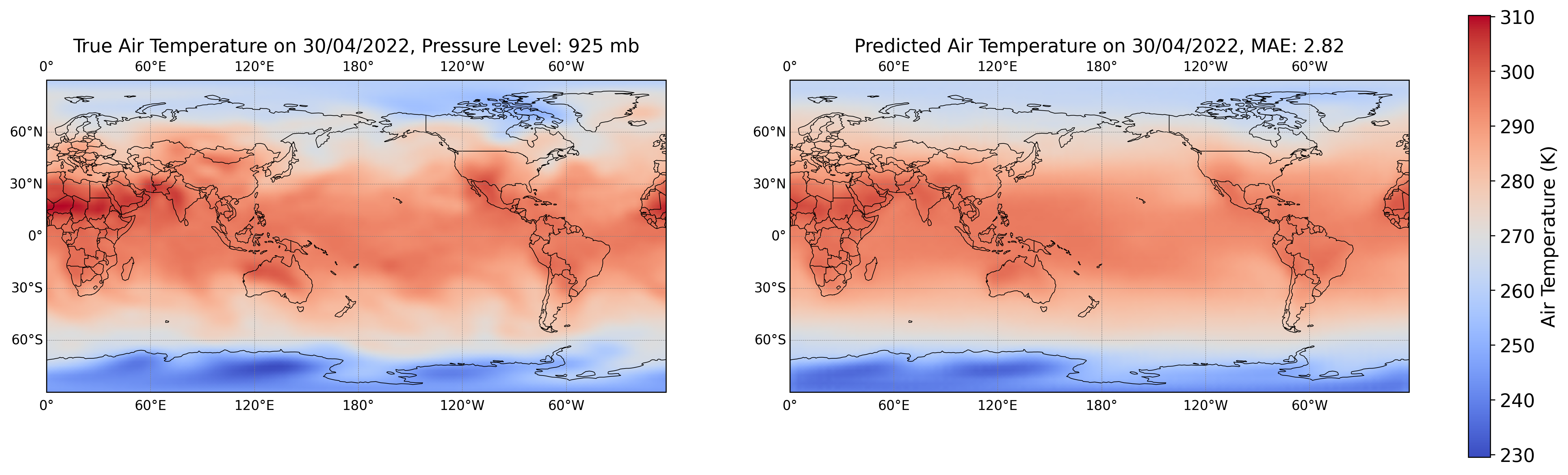}
        \caption{Global air temperature at 925 mb pressure.}
        \label{fig:30_4_2022_Pressure_925}
    \end{subfigure}

    % Subfigure 3
    \begin{subfigure}[b]{1\linewidth}
        \centering
        \includegraphics[width=\linewidth]{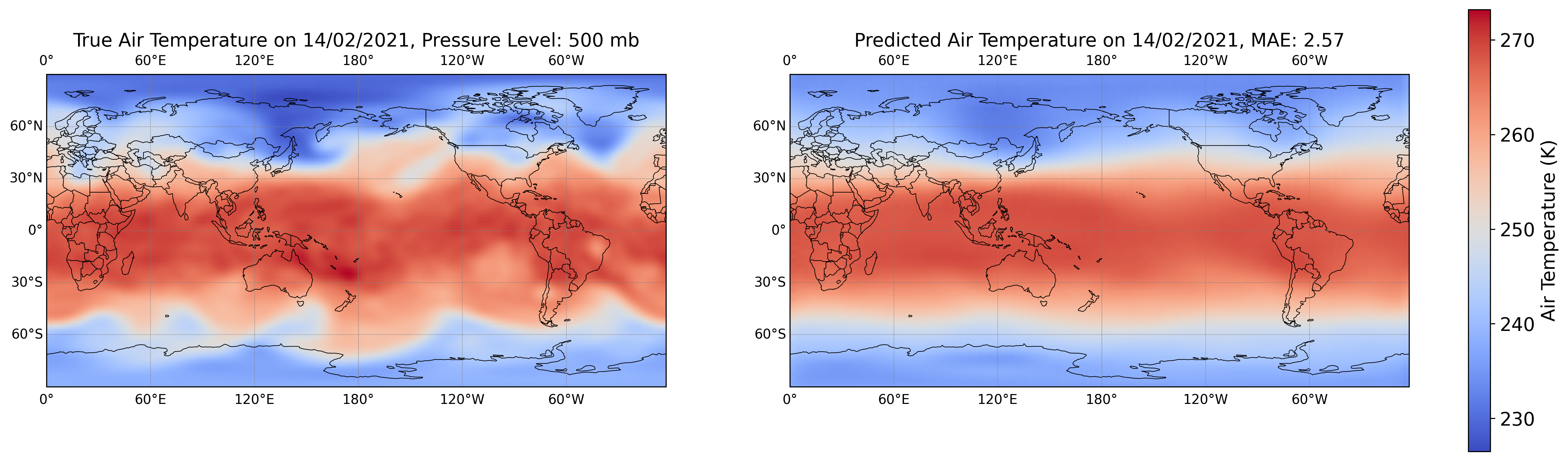}
        \caption{Global air temperature at and 500 mb pressure.}
        \label{fig:14_02_2021_Pressure_500}
    \end{subfigure}

    % Subfigure 4
    \begin{subfigure}[b]{1\linewidth}
        \centering
        \includegraphics[width=\linewidth]{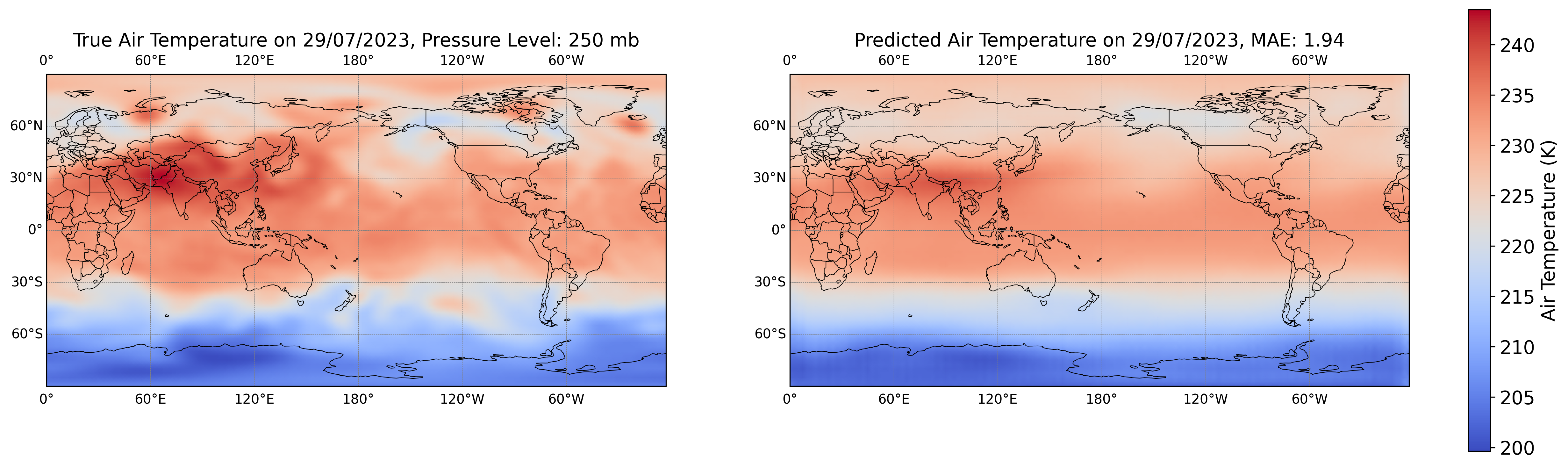}
        \caption{Global air temperature at 250 mb pressure.}
        \label{fig:29_7_2023_Pressure_250}
    \end{subfigure}

    \caption{Testing performance of the TNO for Global Temperature Forecast at various pressure levels.}
    \label{fig:Climate_Pressure_levels}
\end{figure}

Figures \ref{fig:Climate_Pressure1000_Apdx}\subref{fig:19_7_2020_Pressure_1000}, and\ref{fig:Climate_Pressure1000_Apdx}\subref{fig:4_6_2023_Pressure_1000} show comparisons between the ground truth and the predicted air temperatures at various times in the testing period (2019-2023) and at the 1000 mb pressure level. Moreover, Figures \ref{fig:Climate_Pressure_levels}\subref{fig:30_4_2022_Pressure_925}, \ref{fig:Climate_Pressure_levels}\subref{fig:14_02_2021_Pressure_500}, and \ref{fig:Climate_Pressure_levels}\subref{fig:29_7_2023_Pressure_250} show comparisons between the ground truth and the predicted air temperatures at various times and at various atmospheric pressure levels in the testing period (2019-2023).

\newpage
\section{Training and Hyperparameters of  the Geologic Carbon Sequestration Dataset} \label{Tr_Hyp_GCS}

In our third and final example, we use the GCS dataset \cite{Wen2022U-FNOAnFlow} to forecast $\mathrm{CO}_2$ plume migration and pressure buildup in subsurface geological storage. The branch of the TNO processes the scalar and field input variables, the t-branch processes the initial conditions, and the trunk processes the temporal grid, all having a temporal length of $L = 1$. The output of the TNO is the solution at the next three time snapshots ($K = 3$). Moreover, a mask is constructed to account for the varying thicknesses in the various realizations of the dataset. 
Note that the grid invariance property is not an objective in the GCS problem, hence only the temporal dimension is supplied to the trunk, while the spatial grid is introduced as an additional input feature in the branch of each network. This design choice leads to more efficient training without compromising accuracy.

Two TNOs were trained for saturation and pressure buildup for 100 epochs with the Adam optimizer and a weight decay of $1 \times 10^{-4}$ using teacher forcing, followed by 40 epochs of fine-tuning. The initial learning rate is $6 \times 10^{-4}$ and reduces with a \textit{StepLR} scheduler every 2 steps. Both TNOs adopt the same U-Net architecture used in the preceding examples but incorporate a leaky ReLU activation function. In particular, the trunk sub-network employs a tanh activation function, while the output of the t-branch and branch sub-networks is passed through a ReLU activation. Both TNOs are trained with a batch size of 4. In the gas saturation model, each encoder projects its input into a 96-dimensional latent space (i.e., \(p = 96\)), while in the pressure buildup model the latent dimension is set to \(p = 160\), consistent with the mapping defined in equation \ref{eq:lift_to_latent_separate}.

\section{Additional Results for the Geologic Carbon Sequestration Dataset} \label{Results_GCS}
The TNO was trained on the time horizon from zero to 1.8 years only. We classify the TNO performance in the testing dataset into three categories: generalization, simultaneous generalization and short-term extrapolation, and simultaneous generalization and long-term extrapolation. The generalization performance refers to the testing dataset time horizon 0 to 1.8 years for new realizations not seen during training; the simultaneous generalization and short-term extrapolation refers to the testing dataset with a time horizon from 1.8 years up to 7.3 years; the simultaneous generalization and long-term extrapolation refers to the time horizon from 10.4 years onward.

\subsection{Additional Results for Gas Saturation} \label{Results_GCS_GS}

\begin{figure}[!htb]
    \centering
    \includegraphics[width=0.8\linewidth]{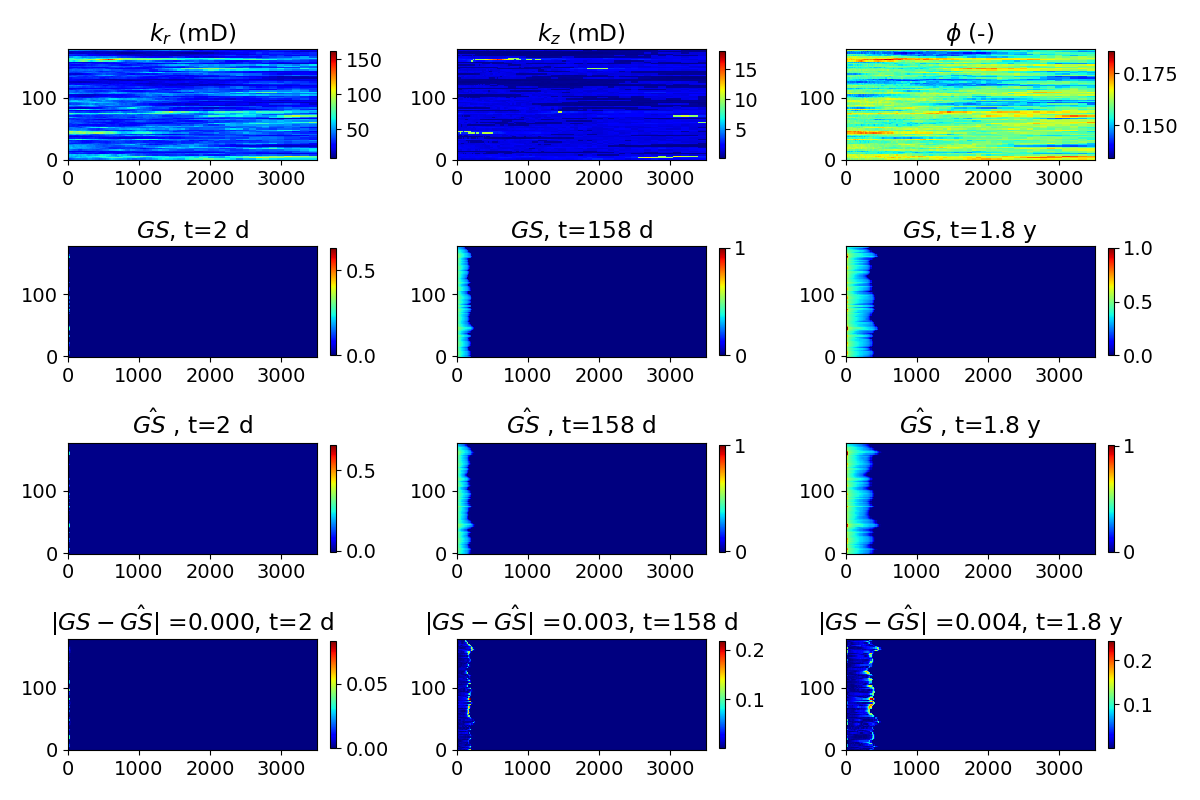}
    \caption{Example 1 of TNO generalization performance on $\mathrm{CO}_2$ saturation. \textit{Row 1}: Input reservoir properties: radial permeability (\(k_r\)), vertical permeability (\(k_z\)), and porosity (\(\phi\)).  
    \textit{Row 2}: Ground truth saturation field.  
    \textit{Row 3}: TNO predicted saturation.  
    \textit{Row 4}: Absolute error. Input parameters: Injection rate: 1.92 MT/yr, temperature: 92.9 °C, initial pressure: 254.7 bar, $S_{wi}$: 0.15, capillary pressure scaling factor: 0.66.}
    \label{fig:Generalization_Test_Picture_1_11_15_num_144}
\end{figure}

\begin{figure}[!htb]
    \centering
    \includegraphics[width=0.8\linewidth]{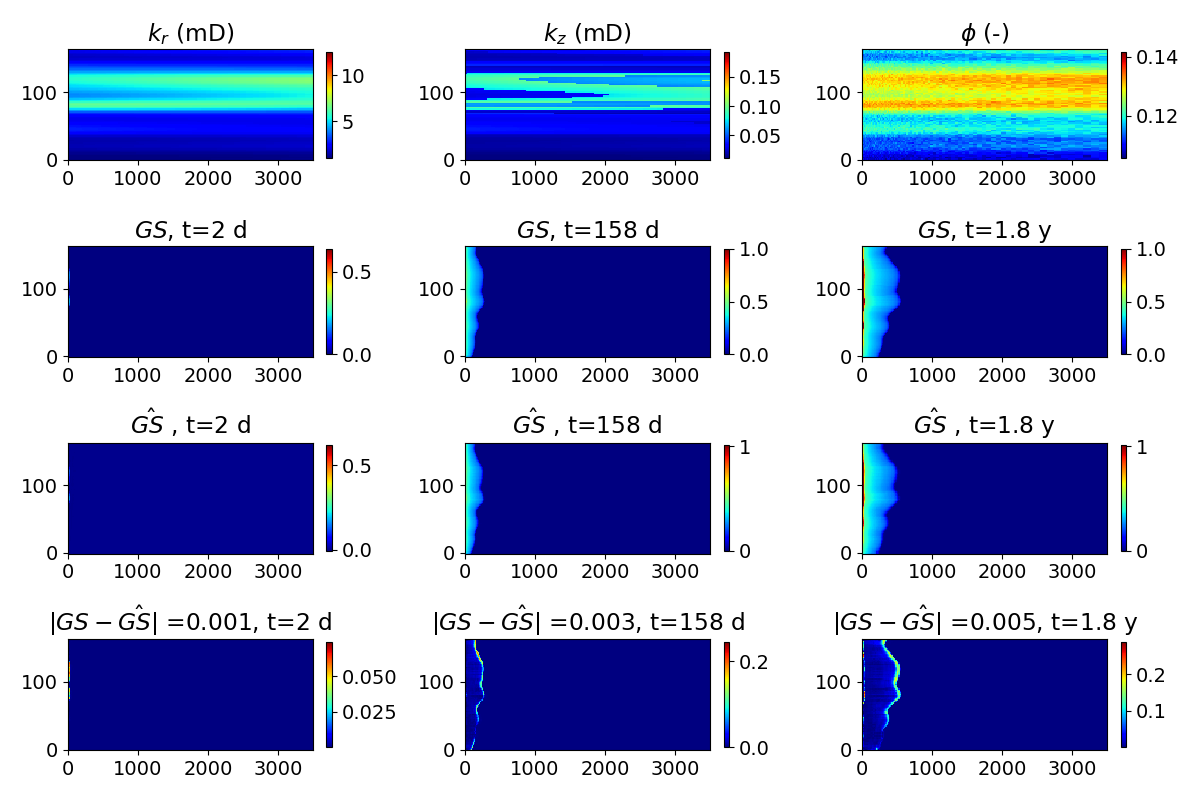}
    \caption{Example 2 of TNO generalization performance on $\mathrm{CO}_2$ saturation. \textit{Row 1}: Input reservoir properties: radial permeability (\(k_r\)), vertical permeability (\(k_z\)), and porosity (\(\phi\)).  
    \textit{Row 2}: Ground truth saturation field.  
    \textit{Row 3}: TNO predicted saturation.  
    \textit{Row 4}: Absolute error. Input parameters: Injection rate: 1.78 MT/yr, temperature: 109.7 °C, initial pressure: 223.8 bar, $S_{wi}$: 0.30, capillary pressure scaling factor: 0.45.}
    \label{fig:Generalization_Test_Picture_1_11_15_num_490}
\end{figure}

\begin{figure}[!htb]
    \centering
    \includegraphics[width=0.8\linewidth]{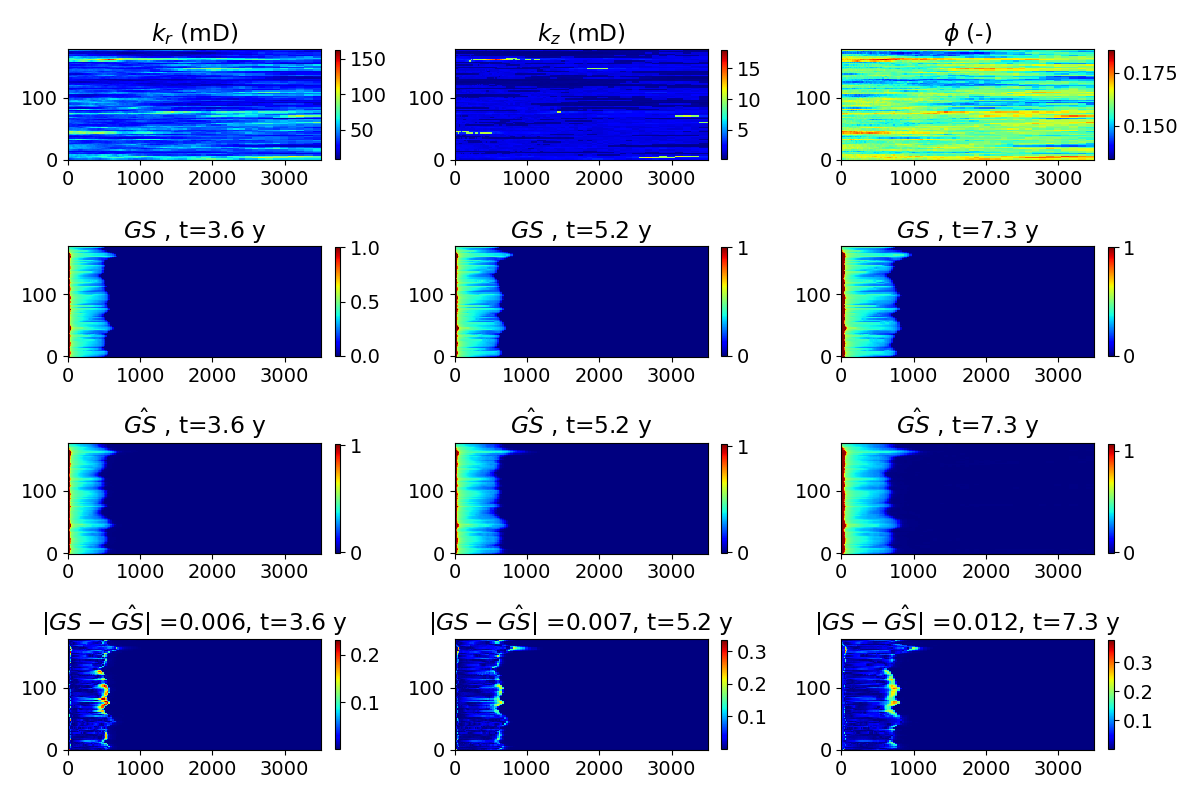}
    \caption{Example 1 of TNO simultaneous generalization and short-term extrapolation performance on $\mathrm{CO}_2$ saturation. \textit{Row 1}: Input reservoir properties: radial permeability (\(k_r\)), vertical permeability (\(k_z\)), and porosity (\(\phi\)).  
    \textit{Row 2}: Ground truth saturation field.  
    \textit{Row 3}: TNO predicted saturation.  
    \textit{Row 4}: Absolute error. Input parameters: Injection rate: 1.92 MT/yr, temperature: 92.9 °C, initial pressure: 254.7 bar, $S_{wi}$: 0.15, capillary pressure scaling factor: 0.66.}
    \label{fig:ShortTermExtra_Test_Picture_17_18_19_num_144}
\end{figure}

\begin{figure}[!htb]
    \centering
    \includegraphics[width=0.8\linewidth]{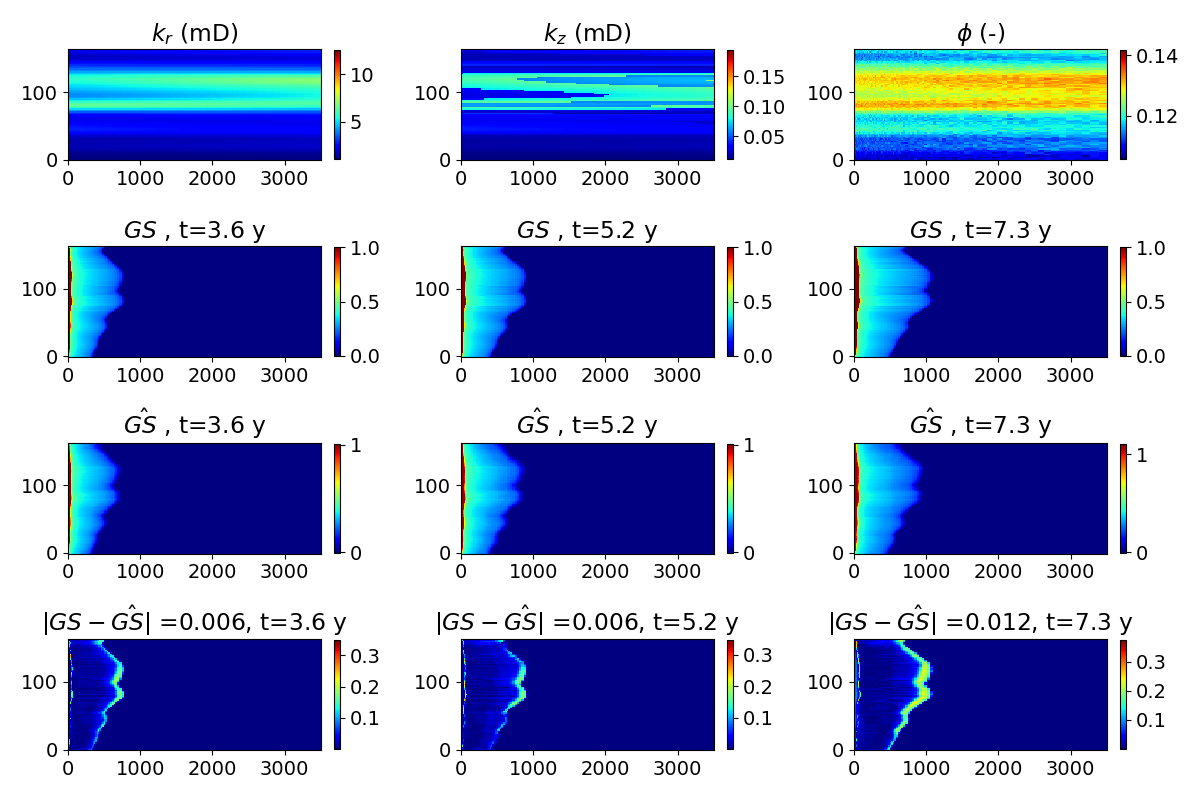}
    \caption{Example 2 of TNO simultaneous generalization and short-term extrapolation performance on $\mathrm{CO}_2$ saturation. \textit{Row 1}: Input reservoir properties: radial permeability (\(k_r\)), vertical permeability (\(k_z\)), and porosity (\(\phi\)).  
    \textit{Row 2}: Ground truth saturation field.  
    \textit{Row 3}: TNO predicted saturation.  
    \textit{Row 4}: Absolute error. Input parameters: Injection rate: 1.78 MT/yr, temperature: 109.7 °C, initial pressure: 223.8 bar, $S_{wi}$: 0.30, capillary pressure scaling factor: 0.45.}
    \label{fig:ShortTermExtra_Test_Picture_17_18_19_num_490}
\end{figure}

\begin{figure}[ht]
    \centering
    \includegraphics[width=0.8\linewidth]{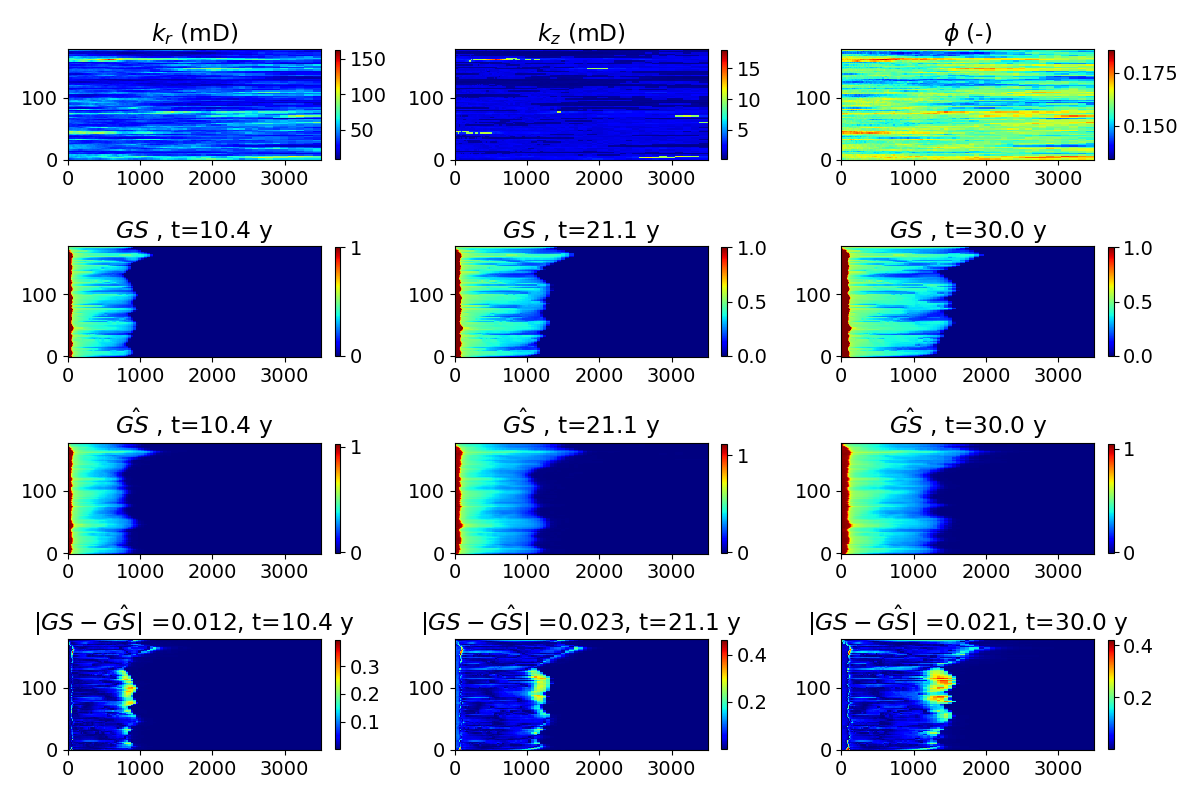}
    \caption{Example 1 of TNO simultaneous generalization and long-term extrapolation performance on $\mathrm{CO}_2$ saturation. \textit{Row 1}: Input reservoir properties: radial permeability (\(k_r\)), vertical permeability (\(k_z\)), and porosity (\(\phi\)).  
    \textit{Row 2}: Ground truth saturation field.  
    \textit{Row 3}: TNO predicted saturation.  
    \textit{Row 4}: Absolute error. Input parameters: Injection rate: 1.92 MT/yr, temperature: 92.9 °C, initial pressure: 254.7 bar, $S_{wi}$: 0.15, capillary pressure scaling factor: 0.66.}
    \label{fig:LongTerm_Extra_Test_Picture_20_21_23_num_144}
\end{figure}

\begin{figure}[!htb]
    \centering
    \includegraphics[width=0.8\linewidth]{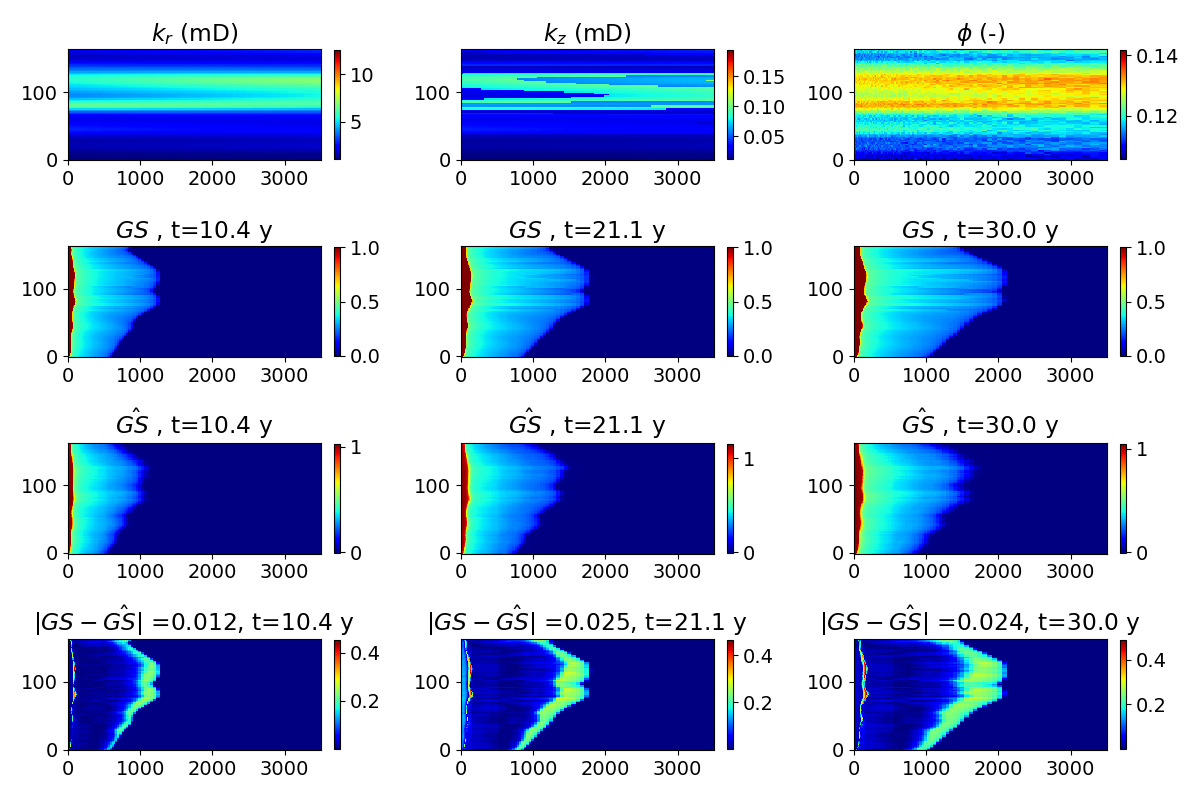}
    \caption{Example 2 of TNO simultaneous generalization and long-term extrapolation performance on $\mathrm{CO}_2$ saturation. \textit{Row 1}: Input reservoir properties: radial permeability (\(k_r\)), vertical permeability (\(k_z\)), and porosity (\(\phi\)).  
    \textit{Row 2}: Ground truth saturation field.  
    \textit{Row 3}: TNO predicted saturation.  
    \textit{Row 4}: Absolute error. Input parameters: Injection rate: 1.78 MT/yr, temperature: 109.7 °C, initial pressure: 223.8 bar, $S_{wi}$: 0.30, capillary pressure scaling factor: 0.45.}
    \label{fig:LongTerm_Extra_Test_Picture_20_21_23_num_490}
\end{figure}

Figures \ref{fig:Generalization_Test_Picture_1_11_15_num_144}, and \ref{fig:Generalization_Test_Picture_1_11_15_num_490} show two examples of TNO generalization performance on gas saturation testing dataset. Figures \ref{fig:ShortTermExtra_Test_Picture_17_18_19_num_144}, and \ref{fig:ShortTermExtra_Test_Picture_17_18_19_num_490} show examples of TNO simultaneous generalization and short-term extrapolation performance on gas saturation testing dataset for the same previous two generalization examples. Figures \ref{fig:LongTerm_Extra_Test_Picture_20_21_23_num_144}, and \ref{fig:LongTerm_Extra_Test_Picture_20_21_23_num_490} show examples of TNO simultaneous generalization and long-term extrapolation performance on gas saturation testing dataset for the same previous two generalization examples.

% simultaneous generalization and short-term extrapolation

% simultaneous generalization and long-term extrapolation

\subsection{Additional Results for Gas Pressure Buildup} \label{Results_GCS_dP}

\begin{figure}[!htb]
    \centering
    \includegraphics[width=0.8\linewidth]{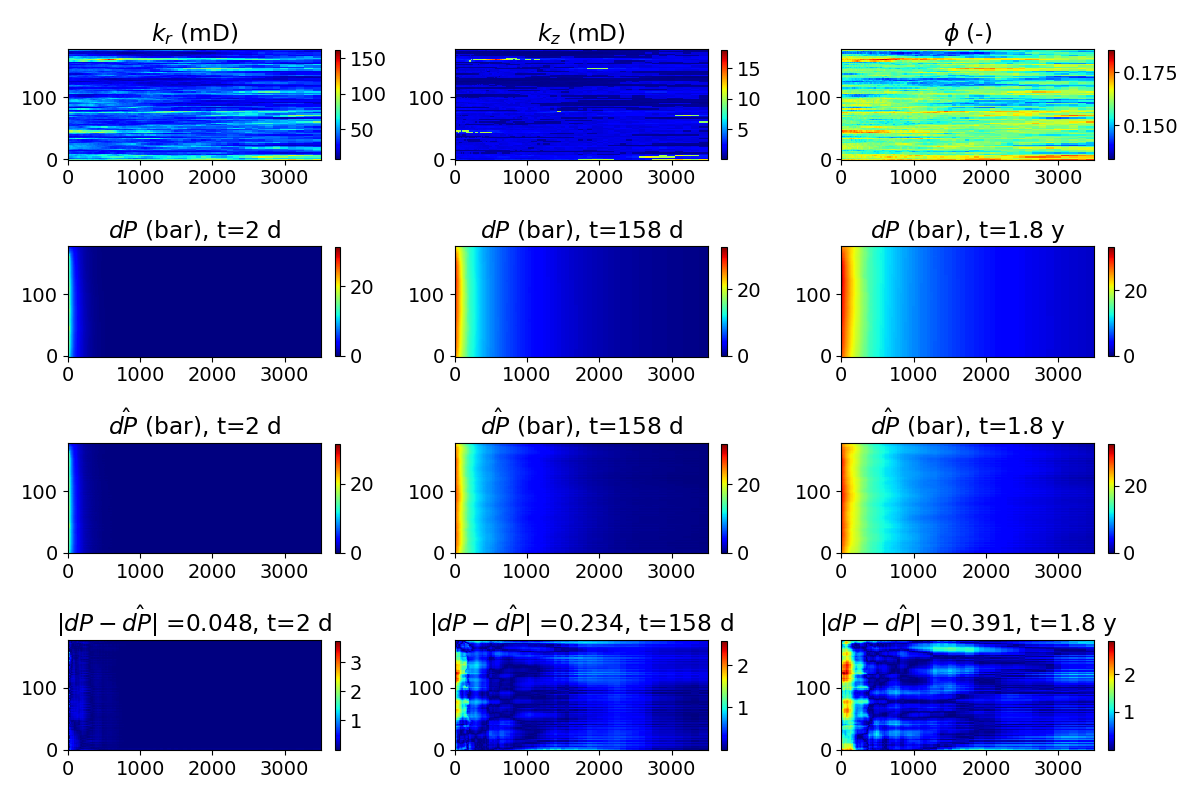}
    \caption{Example 1 of TNO generalization performance on $\mathrm{CO}_2$ pressure buildup. \textit{Row 1}: Input reservoir properties: radial permeability (\(k_r\)), vertical permeability (\(k_z\)), and porosity (\(\phi\)).  
    \textit{Row 2}: Ground truth pressure buildup field.  
    \textit{Row 3}: TNO predicted pressure buildup.  
    \textit{Row 4}: Absolute error. Input parameters: Injection rate: 1.92 MT/yr, temperature: 92.9 °C, initial pressure: 254.7 bar, $S_{wi}$: 0.15, capillary pressure scaling factor: 0.66.}
    \label{fig:dP_Generalization_Test_Picture_1_11_15_num_144}
\end{figure}

\begin{figure}[!htb]
    \centering
    \includegraphics[width=0.8\linewidth]{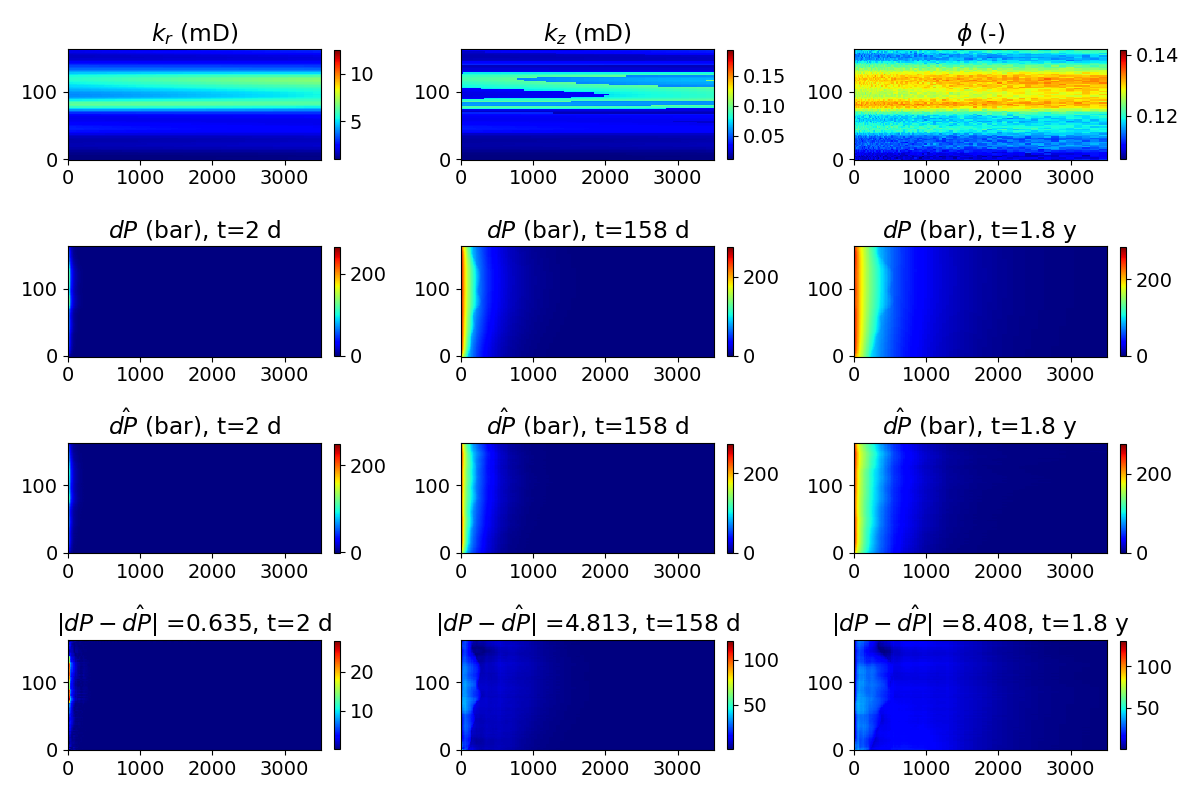}
    \caption{Example 2 of TNO generalization performance on $\mathrm{CO}_2$ pressure buildup. \textit{Row 1}: Input reservoir properties: radial permeability (\(k_r\)), vertical permeability (\(k_z\)), and porosity (\(\phi\)).  
    \textit{Row 2}: Ground truth pressure buildup field.  
    \textit{Row 3}: TNO predicted pressure buildup.  
    \textit{Row 4}: Absolute error. Input parameters: Injection rate: 1.78 MT/yr, temperature: 109.7 °C, initial pressure: 223.8 bar, $S_{wi}$: 0.30, capillary pressure scaling factor: 0.45.}
    \label{fig:dP_Generalization_Test_Picture_1_11_15_num_490}
\end{figure}

\begin{figure}[!htb]
    \centering
    \includegraphics[width=0.75\linewidth]{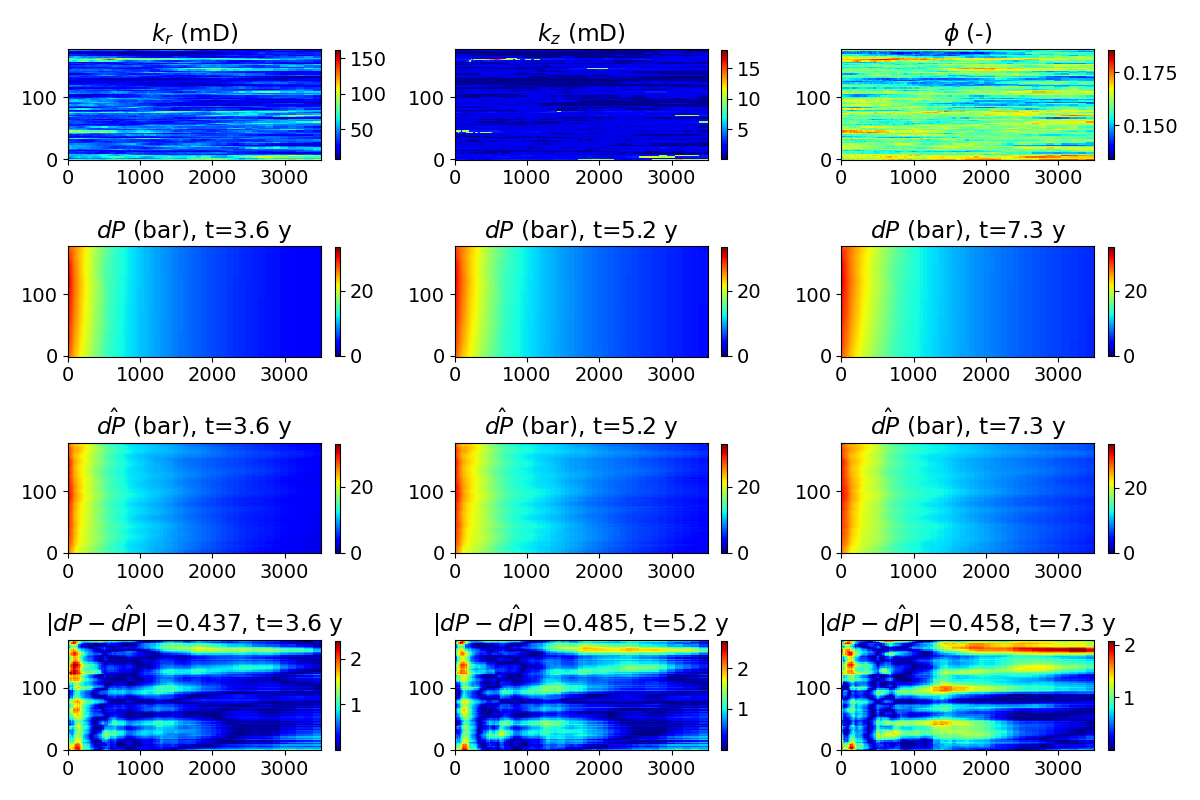}
    \caption{Example 1 of TNO simultaneous generalization and short-term extrapolation performance on $\mathrm{CO}_2$ pressure buildup. \textit{Row 1}: Input reservoir properties: radial permeability (\(k_r\)), vertical permeability (\(k_z\)), and porosity (\(\phi\)).  
    \textit{Row 2}: Ground truth pressure buildup field.  
    \textit{Row 3}: TNO predicted pressure buildup.  
    \textit{Row 4}: Absolute error. Input parameters: Injection rate: 1.92 MT/yr, temperature: 92.9 °C, initial pressure: 254.7 bar, $S_{wi}$: 0.15, capillary pressure scaling factor: 0.66.}
    \label{fig:dP_ShortTermExtra_Test_Picture_17_18_19_num_144}
\end{figure}

\begin{figure}[!htb]
    \centering
    \includegraphics[width=0.75\linewidth]{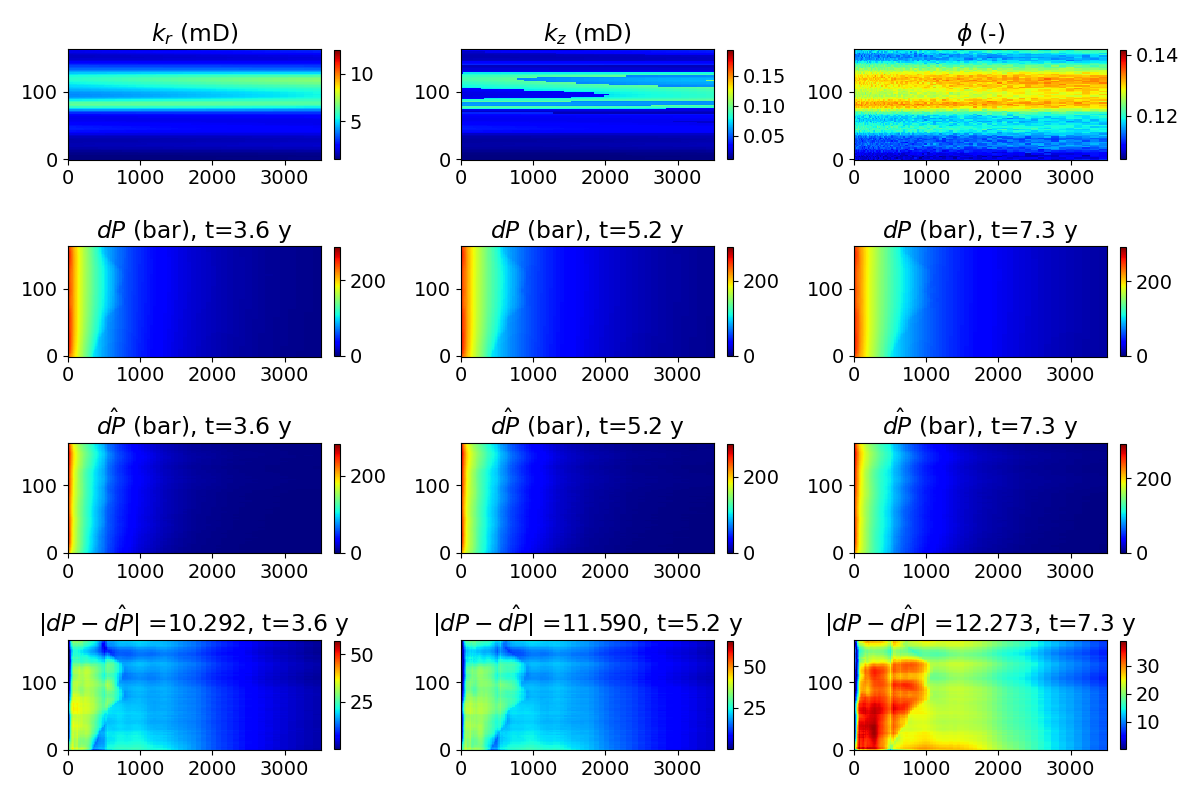}
    \caption{Example 2 of TNO simultaneous generalization and short-term extrapolation performance on $\mathrm{CO}_2$ pressure buildup. \textit{Row 1}: Input reservoir properties: radial permeability (\(k_r\)), vertical permeability (\(k_z\)), and porosity (\(\phi\)).  
    \textit{Row 2}: Ground truth pressure buildup field.  
    \textit{Row 3}: TNO predicted pressure buildup.  
    \textit{Row 4}: Absolute error. Input parameters: Injection rate: 1.78 MT/yr, temperature: 109.7 °C, initial pressure: 223.8 bar, $S_{wi}$: 0.30, capillary pressure scaling factor: 0.45.}
    \label{fig:dP_ShortTermExtra_Test_Picture_17_18_19_num_490}
\end{figure}

\begin{figure}[ht]
    \centering
    \includegraphics[width=0.75\linewidth]{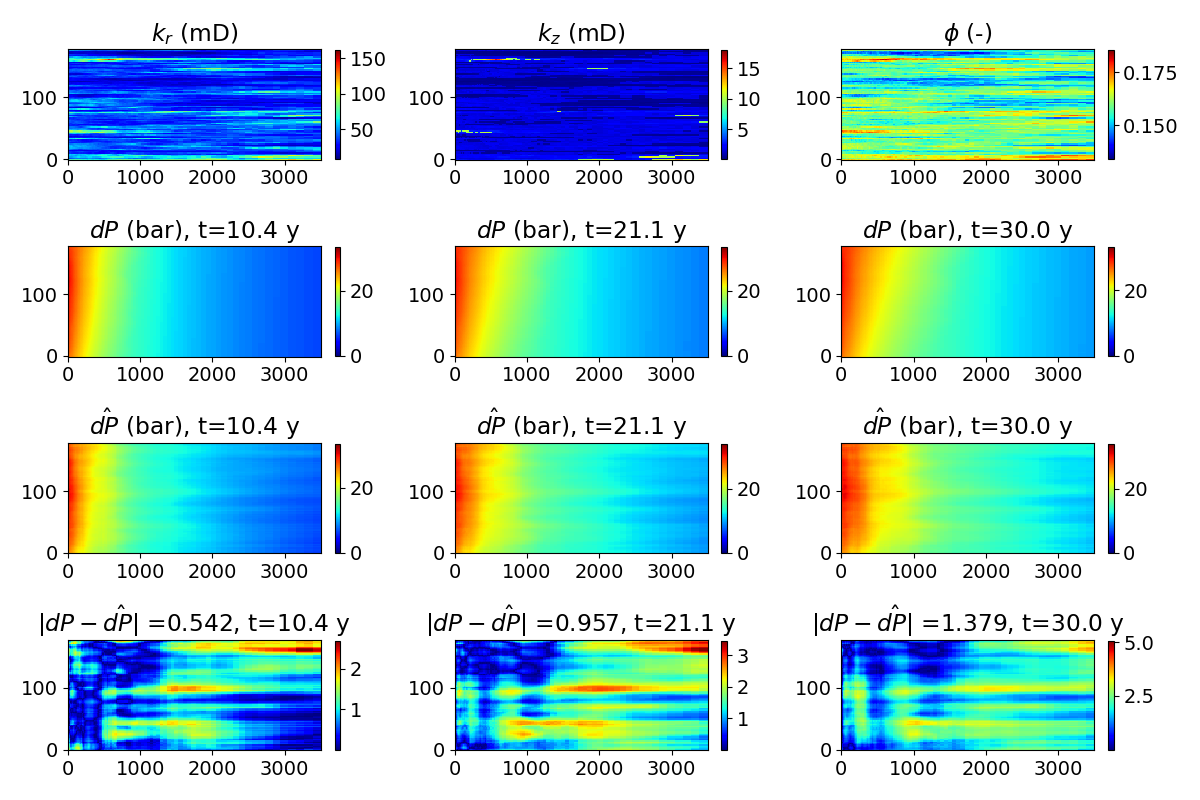}
    \caption{Example 1 of TNO simultaneous generalization and long-term extrapolation performance on $\mathrm{CO}_2$ pressure buildup. \textit{Row 1}: Input reservoir properties: radial permeability (\(k_r\)), vertical permeability (\(k_z\)), and porosity (\(\phi\)).  
    \textit{Row 2}: Ground truth pressure buildup field.  
    \textit{Row 3}: TNO predicted pressure buildup.  
    \textit{Row 4}: Absolute error. Input parameters: Injection rate: 1.92 MT/yr, temperature: 92.9 °C, initial pressure: 254.7 bar, $S_{wi}$: 0.15, capillary pressure scaling factor: 0.66.}
    \label{fig:dP_LongTerm_Extra_Test_Picture_20_21_23_num_144}
\end{figure}

\begin{figure}[!htb]
    \centering
    \includegraphics[width=0.75\linewidth]{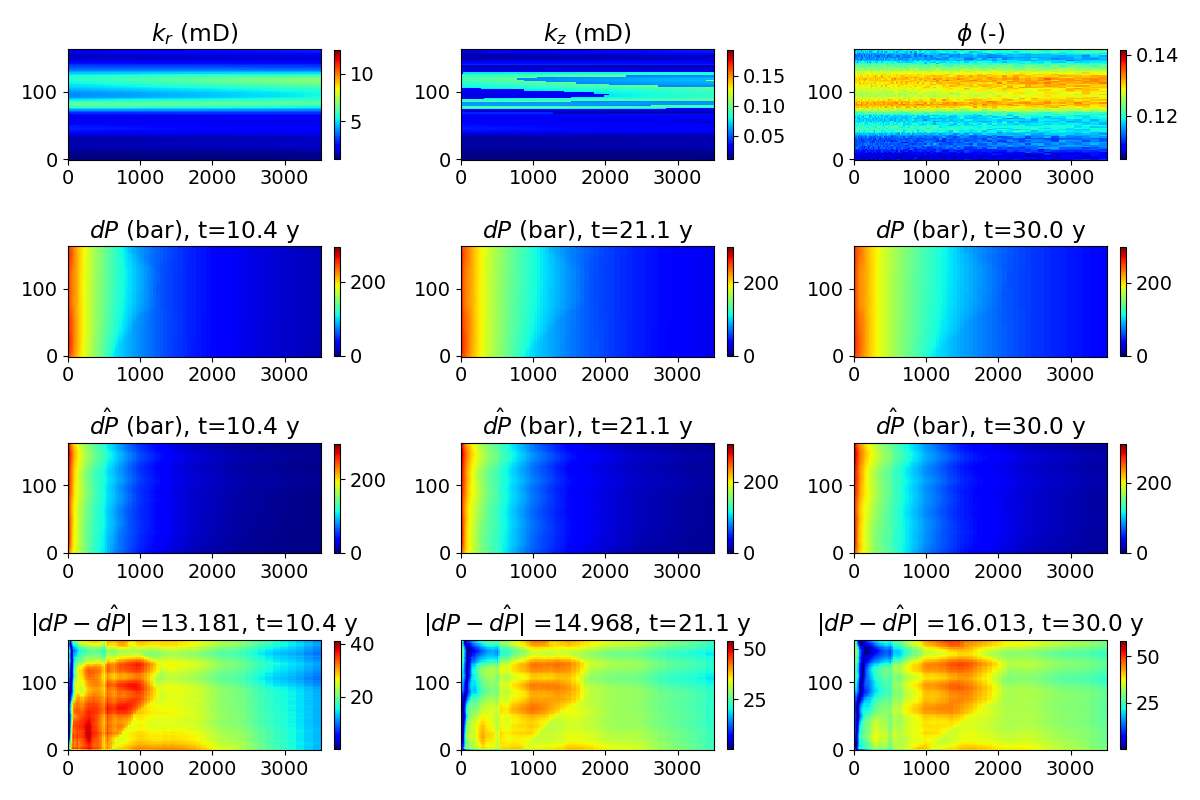}
    \caption{Example 2 of TNO simultaneous generalization and long-term extrapolation performance on $\mathrm{CO}_2$ pressure buildup. \textit{Row 1}: Input reservoir properties: radial permeability (\(k_r\)), vertical permeability (\(k_z\)), and porosity (\(\phi\)).  
    \textit{Row 2}: Ground truth pressure buildup field.  
    \textit{Row 3}: TNO predicted pressure buildup.  
    \textit{Row 4}: Absolute error. Input parameters: Injection rate: 1.78 MT/yr, temperature: 109.7 °C, initial pressure: 223.8 bar, $S_{wi}$: 0.30, capillary pressure scaling factor: 0.45.}
    \label{fig:dP_LongTerm_Extra_Test_Picture_20_21_23_num_490}
\end{figure}

Figures \ref{fig:dP_Generalization_Test_Picture_1_11_15_num_144}, and \ref{fig:dP_Generalization_Test_Picture_1_11_15_num_490} show two examples of TNO generalization performance on $\mathrm{CO}_2$ pressure buildup testing dataset. Figures \ref{fig:dP_ShortTermExtra_Test_Picture_17_18_19_num_144}, and \ref{fig:dP_ShortTermExtra_Test_Picture_17_18_19_num_490} show examples of TNO simultaneous generalization and short-term extrapolation performance on $\mathrm{CO}_2$ pressure buildup testing dataset for the same previous two generalization examples. Figures \ref{fig:dP_LongTerm_Extra_Test_Picture_20_21_23_num_144}, and \ref{fig:dP_LongTerm_Extra_Test_Picture_20_21_23_num_490} show examples of TNO simultaneous generalization and long-term extrapolation performance on $\mathrm{CO}_2$ pressure buildup testing dataset for the same previous two generalization examples.

% simultaneous generalization and short-term extrapolation

% simultaneous generalization and long-term extrapolation

\end{document}